\documentclass[twoside,11pt]{article}

\usepackage{jmlr2e}

\usepackage{algpseudocode}
\usepackage{algorithm}
\usepackage{multirow}
\usepackage{hyperref}
\hypersetup{
    pdfnewwindow=true,     % links in new window
    colorlinks=true,     % false: boxed links; true: colored links
    linkcolor=blue,     % color of internal links (change box color with linkbordercolor)
    citecolor=blue,     % color of links to bibliography
    filecolor=magenta,     % color of file links
    urlcolor=cyan     % color of external links
}

\usepackage{amsmath} %,amsthm,amssymb,amsfonts}
\usepackage[capitalize]{cleveref}
\usepackage[normalem]{ulem}
\usepackage{cancel}
\usepackage{mathtools}
\usepackage{bbm,dsfont}
\usepackage{enumitem}
\usepackage{xspace}
\usepackage{graphicx}

\newcommand{\hL}{\hat{L}}
\newcommand{\ce}{\text{x-e}}

\newcommand{\KL}{\operatorname{KL}} % KL divergence - general
\newcommand{\kl}{\operatorname{kl}} % KL divergence - binary
\newcommand{\E}{\operatorname{\mathbb{E}}} % expectation
\newcommand{\EE}{\operatorname{\mathbb{E}}} % expectation
\newcommand{\R}{\mathbb{R}}  % reals
\newcommand*{\quotientspace}[2]% quotient space displayed
{\ensuremath{\raisebox{.25ex}{\ensuremath{#1}} \hspace{-.35ex} \raisebox{-.25ex}{/} 
\hspace{-.05ex} \raisebox{-.85ex}{\ensuremath{#2}}} \hspace{.2ex} }
\newcommand*{\textquotientspace}[2]% quotient space inside text
{\ensuremath{\raisebox{.05ex}{\ensuremath{#1}} \hspace{-.30ex} \raisebox{-.15ex}{/} 
\hspace{.10ex} \raisebox{-.55ex}{\ensuremath{#2}}}}

\newcommand{\upsum}% upper sum sign 
{\ensuremath \mkern5.4mu \overline{\vphantom{S}\mkern8mu} \mkern-11mu S}
\newcommand{\lowsum}% lower sum sign 
{\ensuremath \mkern3.4mu \underline{\vphantom{S}\mkern8mu} \mkern-9mu S}
\newcommand{\upint}% upper integral sign 
{\ensuremath \mkern10.4mu \overline{\vphantom{\int}\mkern7mu} \mkern-21mu\int}
\newcommand{\upintwdomain}[1]% upper integral sign w/domain
{\ensuremath \mkern10.4mu \overline{\vphantom{\int}\mkern7mu} \mkern-21mu\int_{#1}}
\newcommand{\textupint}% upper integral sign inside text
{\textstyle \mkern5mu \overline{\vphantom{\int}\mkern6mu} \mkern-14mu\intop\mkern0mu}
\newcommand{\textupintwdomain}[1]% upper integral sign inside text w/domain
{\textstyle \mkern5mu \overline{\vphantom{\int}\mkern6mu} \mkern-14mu\intop_{#1}}
\newcommand{\lowint}% lower integral sign
{\ensuremath \underline{\vphantom{\int}\mkern7mu} \mkern-9mu\int}
\newcommand{\lowintwdomain}[1]% lower integral sign w/domain
{\ensuremath \underline{\vphantom{\int}\mkern7mu} \mkern-9mu\int_{#1}}
\newcommand{\textlowint}% lower integral sign inside text
{\textstyle \mkern3mu \underline{\vphantom{\int}\mkern6mu} \mkern-8.5mu\intop}
\newcommand{\textlowintwdomain}[1]% lower integral sign inside text w/domain
{\textstyle \mkern3mu \underline{\vphantom{\int}\mkern6mu} \mkern-8.5mu\intop_{#1}}

\def\ddefloop#1{\ifx\ddefloop#1\else\ddef{#1}\expandafter\ddefloop\fi}

\def\ddef#1{\expandafter\def\csname c#1\endcsname{\ensuremath{\mathcal{#1}}}}
\ddefloop ABCDEFGHIJKLMNOPQRSTUVWXYZ\ddefloop

\def\ddef#1{\expandafter\def\csname bb#1\endcsname{\ensuremath{\mathbb{#1}}}}
\ddefloop ABCDEFGHIJKLMNOPQRSTUVWXYZ\ddefloop

\def\ddef#1{\expandafter\def\csname fr#1\endcsname{\ensuremath{\mathfrak{#1}}}}
\ddefloop ABCDEFGHIJKLMNOPQRSTUVWXYZabcdefghijklmnopqrstuvwxyz\ddefloop

\def\ddef#1{\expandafter\def\csname ss#1\endcsname{\ensuremath{\mathsf{#1}}}}
\ddefloop ABCDEFGHIJKLMNOPQRSTUVWXYZabcdefghijklmnopqrstuvwxyz\ddefloop

\def\ddef#1{\expandafter\def\csname bf#1\endcsname{\ensuremath{\mathbf{#1}}}}
\ddefloop ABCDEFGHIJKLMNOPQRSTUVWXYZabcdefghijklmnopqrstuvwxyz\ddefloop

\def\ddef#1{\expandafter\def\csname v#1\endcsname{\ensuremath{\boldsymbol{#1}}}}
\ddefloop ABCDEFGHIJKLMNOPQRSTUVWXYZabcdefghijklmnopqrstuvwxyz\ddefloop

\def\ddef#1{\expandafter\def\csname v#1\endcsname{\ensuremath{\boldsymbol{\csname #1\endcsname}}}}
\ddefloop {alpha}{beta}{gamma}{delta}{epsilon}{varepsilon}{zeta}{eta}{theta}{vartheta}{iota}{kappa}{lambda}{mu}{nu}{xi}{pi}{varpi}{rho}{varrho}{sigma}{varsigma}{tau}{upsilon}{phi}{varphi}{chi}{psi}{omega}{Gamma}{Delta}{Theta}{Lambda}{Xi}{Pi}{Sigma}{varSigma}{Upsilon}{Phi}{Psi}{Omega}\ddefloop
\usepackage{lastpage}
\jmlrheading{22}{2021}{1-\pageref{LastPage}}{8/20; Revised
5/21}{8/21}{20-879}
{Mar{\'i}a P{\'e}rez-Ortiz, Omar Rivasplata, John Shawe-Taylor and Csaba Szepesv\'ari}

\ShortHeadings{Tighter Risk Certificates for Neural Networks}
{P\'erez-Ortiz, Rivasplata, Shawe-Taylor and Szepesv\'ari}

\firstpageno{1}

\begin{document}

\title{Tighter Risk Certificates for Neural Networks}

\author{\name Mar\'ia P\'erez-Ortiz
       \email maria.perez@ucl.ac.uk \\
       \addr AI Centre, University College London (UK)
       %\\
       %Country
       \AND
       \name Omar Rivasplata
       \email o.rivasplata@cs.ucl.ac.uk \\
       \addr AI Centre, University College London (UK)
       %\\
       %Country
       \AND
       \name John Shawe-Taylor
       \email j.shawe-taylor@ucl.ac.uk \\
       \addr AI Centre, University College London (UK)
       %\\
       %Country
       \AND
       \name Csaba Szepesv\'ari 
       \email szepi@google.com\\
       \addr DeepMind Edmonton (Canada)
       %\\
       %Country
       }

\editor{Arnak Dalalyan}

\maketitle

\begin{abstract}%
This paper presents an empirical study regarding
training probabilistic neural networks using training objectives derived from PAC-Bayes bounds.
In the context of probabilistic neural networks, the output of training is a probability distribution over network weights.
We present two training objectives, used here for the first time in connection with training neural networks. 
These two training objectives are derived from tight PAC-Bayes bounds. 
We also re-implement a previously used training objective based on a classical PAC-Bayes bound, to compare the properties of the predictors learned using the different training objectives.
We compute risk certificates for the learnt predictors, based on part of the data used to learn the predictors.
We further experiment with different types of priors on the weights (both data-free and data-dependent priors) and neural network architectures.
Our experiments on MNIST and CIFAR-10 show that
our training methods produce competitive test set errors 
and non-vacuous risk bounds with much tighter values than previous results in the literature, showing promise not only to guide the learning algorithm through bounding the risk but also for model selection. 
These observations suggest that the methods studied here might be good candidates for self-certified learning, 
% in the sense of certifying the risk on any unseen data (from the same distribution as the training data) with a numerical certificate that is evaluated on the same data set that was used to learn the predictor.
in the sense of using the whole data set for learning a predictor and certifying its risk on any unseen data (from the same distribution as the training data) potentially without the need for holding out test data.
\end{abstract}


\section{Introduction}
\label{s:intro}

In a probabilistic neural network, the result of the training process is a % 
distribution over network weights, rather than simply fixed weights.
Several prediction schemes can be devised based on a probability distribution over weights.
For instance, one may use a randomised predictor, where each prediction is done by randomly sampling the weights from the data-dependent distribution obtained as the result of the training process. Another prediction rule consists of predicting always with the mean of the learned distribution. Yet another prediction rule is the ensemble predictor based on integrating the predictions of all possible parameter settings, weighted according to the learned distribution.

In this paper we experiment with probabilistic neural networks from a PAC-Bayes approach.
We name `PAC-Bayes with Backprop' (PBB) the family of (probabilistic) neural network training methods derived from PAC-Bayes bounds and %  
optimised through stochastic gradient descent.
The work reported here is the result of our empirical studies undertaken to % 
investigate three PBB training objectives. For reference, they are the functions $f_{\mathrm{quad}}$, $f_{\mathrm{lambda}}$ and $f_{\mathrm{classic}}$, shown respectively in Eq.~\eqref{eq:obj-quad}, Eq.~\eqref{eq:obj-lambda} and Eq.~\eqref{eq:obj-classic} below.
These objectives are based on PAC-Bayes bounds with similar names, which are relaxations of the PAC-Bayes relative entropy bound \citep{LangfordSeeger2001}, also known %
as the PAC-Bayes-kl bound in the literature. 
The classic PAC-Bayes bound, from which $f_{\mathrm{classic}}$ is derived, is that of \cite{McAllester1999}, but we use it with the improved dependence on the number of training patterns as clarified by \cite{maurer2004note}.
The PAC-Bayes-lambda bound is that of \cite{thiemann-etal2017}.
The PAC-Bayes-quadratic bound, from which $f_{\mathrm{quad}}$ is derived, was first introduced by us in the preprint \citet{rivasplata2019pbb}.
Importantly, our work shows tightness of the numerical certificates on the error of the randomised classifiers generated by these training methods.
In each case, the computed certificate is valid on unseen examples (from the same data distribution as the training data), and is evaluated using (part of) the data set that was used to learn the predictor for which the certificate is valid.
These properties make our work a first %
example of \emph{self-certified learning}, which proposes to use the whole data set for learning a predictor 
% (
and certifying its risk on unseen data, without the need for data splitting protocols both for testing and model selection.

Our line of research owes credit to previous works that have trained a probabilistic neural network by minimising a PAC-Bayes bound, or used a PAC-Bayes bound to give risk certificates for trained neural networks.
\citet{LangfordCaruana2001} developed a method to train a probability distribution over neural network weights by randomising the weights with Gaussian noise (adjusted in a data-dependent way via a sensitivity analysis), and computed an upper bound on the error using the PAC-Bayes-kl bound.%
\footnote{Inversion of the PAC-Bayes-kl bound (we explain this in \cref{s:kl_invert}) gives a certificate (upper bound) on the risk of the randomised predictor, in terms of its empirical error and other quantities. The empirical error term is evaluated indirectly by Monte Carlo sampling, and a bound on the tail of the Monte Carlo evaluation \citep[Theorem 2.5]{LangfordCaruana2001} is combined with the PAC-Bayes-kl bound to give a numerical risk certificate that holds with high probability over data and Monte Carlo samples.} 
They also suggested that PAC-Bayes bounds might be fruitful for computing non-vacuous generalisation bounds for neural nets.
\citet{dziugaite2017computing} used a training objective (essentially equivalent to $f_{\mathrm{classic}}$) based on a relaxation of the PAC-Bayes-kl bound. They optimised this objective using stochastic gradient descent (SGD), and computed a confidence bound on the error of the randomised classifier following the same approach that \citet{LangfordCaruana2001} used to compute their error bound.
\citet{dziugaite2018data} developed a two-stage method, 
which in the first stage trains a prior mean by empirical risk minimisation via stochastic gradient Langevin dynamics  \citep{WellingTeh2011}, and in the second stage re-uses the same 
data used for the prior in order to train a posterior Gibbs distribution over weights; they also evaluate a relaxation of the PAC-Bayes-kl bound, based on ideas from differential privacy \citep{dwork-etal2015gen,dwork-etal2015preserving}, which accounts for the data re-use.

In this paper we report experiments on MINIST and CIFAR-10 with the three training objectives mentioned above. 
We used by default the randomised predictor scheme (also called the `stochastic predictor' in the PAC-Bayes literature), justified by the fact that PAC-Bayes bounds give high-confidence guarantees on the expected loss of the randomised predictor. 
Since training is based on a surrogate loss function, optimising a PBB objective gives a high-confidence guarantee on the randomised predictor's risk under the surrogate loss. 
Accordingly, to obtain guarantees that are valid for the classification error (i.e. the zero-one loss), we separately evaluate a confidence bound for the error based on part of the data that was used to learn the randomised predictor (following the procedure that was used by \cite{LangfordCaruana2001} and \cite{dziugaite2017computing}).
For comparison we also report test set error for the randomised predictor, and for the other two predictor schemes described above, namely, the posterior mean and the ensemble predictors.

Our work took inspiration from \citet{blundell2015weight}, whose results showed that randomised weights achieve competitive test set errors;
and from \citet{dziugaite2017computing,dziugaite2018data}, whose results gave randomised neural network classifiers with reasonable test set errors and, more importantly, non-vacuous risk bound values. 
Our experiments %
show that PBB training objectives can 
(a) achieve competitive test set errors (e.g. comparable to \citet{blundell2015weight} and empirical risk minimisation), while also
(b) deliver risk certificates with reasonably tight values. 
Our results show as well a significant improvement over 
those of \citet{dziugaite2017computing,dziugaite2018data}: we further close the gap between the numerical risk certificate (bound value) and the risk estimate (test set error rate). 
As we argue below, this improvement comes from the tightness of the PAC-Bayes bounds we used, which is established analytically and corroborated by our experiments on MNIST and CIFAR-10 with deep fully connected networks and convolutional neural networks.

Regarding the tightness,
the training objective of \citet{dziugaite2017computing} (which in our notation takes essentially the form of $f_{\mathrm{classic}}$ shown in Eq.~\eqref{eq:obj-classic} below) has the disadvantage of being sub-optimal in the regime of small losses. 
This is because their objective is a relaxation of the PAC-Bayes-kl bound via an inequality that is loose in this regime. The looseness was the price paid for having a computable objective. 
Note that small losses is precisely the regime of interest in neural network training (although the true loss being small is data set and architecture dependent).
By contrast, our proposed training % 
objectives ($f_{\mathrm{quad}}$ and $f_{\mathrm{lambda}}$ in Eq.~\eqref{eq:obj-quad} and Eq.~\eqref{eq:obj-lambda} below) are based on relaxing the PAC-Bayes-kl bound by an inequality that is tighter in this same regime of small losses, which is one of the reasons explaining our tighter risk certificates in MNIST (not for CIFAR-10, which could be explained by the large empirical loss obtained at the end of the optimisation). Interestingly, our own re-implementation of $f_{\mathrm{classic}}$ also gave improved results compared to the results of Dziugaite and Roy, which suggests that besides the training objectives we used, also the training strategies we used are responsible for the improvements.

A clear advantage of PAC-Bayes with Backprop (PBB) methods is being an instance of self-certified\footnote{We say that a learning method is \emph{self-certified} if it uses all the available data in order to simultaneously output a predictor and a reasonably tight risk certificate that is valid on unseen data.} learning: When training probabilistic neural nets by PBB methods the output % 
is not just a predictor but simultaneously a \emph{tight risk certificate} that guarantees the quality of predictions on unseen examples.
The value of self-certified learning algorithms (cf. \citealp{Freund1998}) is in the possibility of using of all the available data to achieve both goals (learning a predictor and certifying its risk) simultaneously, thus making efficient use of the available data.
Note that risk certificates \emph{per se} will not impress until their reported values  match or closely follow the classification error rates evaluated on a test set, so that the risk certificate is % 
informative of 
the out-of-sample error.
This is where our work makes a significant contribution, since our PBB training methods lead to risk certificates for neural nets with much tighter values than previous works in the literature.
Once again, the %
solution found by our
learning procedure
comes together with a high-confidence guarantee that certifies its risk under the surrogate training loss, and to obtain a high-confidence guarantee for the classification error (zero-one loss) we evaluate \emph{post training} a risk bound.
A more ambitious goal would be to establish calibration\footnote{This is akin to results on calibration of the surrogate hinge loss, cf. \citet{SteinwartChristmann2008SVMs}.} of the surrogate cross-entropy loss, so then minimising it would guarantee minimal classification error.%

We would like to highlight the elegant simplicity of the methods presented here: Our results are achieved 
i) with priors learnt through empirical risk minimisation of the surrogate loss on a subset of the data set (which does not overlap with the data used for computing the risk certificate for the %
randomised predictor,
thus in line with classical PAC-Bayes priors) and 
ii) via classical SGD optimisation. 
In contrast, \citet{dziugaite2018data}  trained a special type of data-dependent PAC-Bayes prior on the whole data set using SGLD optimisation. They justified this procedure arguing that the limit distribution of SGLD satisfies the differential privacy property (but a finite-time guarantee was missing), and relaxed the PAC-Bayes-kl bound with a correction term based on the concept of max-information\footnote{\cite{dwork-etal2015gen,dwork-etal2015preserving} proposed this concept in the context of adaptive data analysis.} to account for using the same data to train the prior mean and to evaluate the  bound.
Furthermore, our methods do not involve tampering with the training objective, as opposed to \citet{blundell2015weight}, who used a ``KL attenuating trick'' 
by inserting a tunable parameter as a factor of the Kullback-Leibler (KL) divergence in their objective. 
Our work highlights the point that it is worthwhile studying simple methods, not just to understand their scope or for the sake of having a more controlled experimental setup, but also to more accurately assess the real value added by the `extras' of the more complex methods.%

\smallskip

\noindent\textit{Our Contributions:}
\begin{enumerate}[leftmargin=*,topsep=2pt,itemsep=2pt]
    \item We rigorously study and illustrate `PAC-Bayes with Backprop' (PBB), a generic strategy to derive (probabilistic) neural network training methods from PAC-Bayes bounds.
    \item We propose ---and experiment with--- two new PBB training objectives: 
    one derived from % our novel 
    the PAC-Bayes-quadratic bound of \citet{rivasplata2019pbb}, and
    one derived from the PAC-Bayes-lambda bound of \citet{thiemann-etal2017}.
    \item We also re-implement the training objective based on the classic PAC-Bayes bound that was used by Dziugaite and Roy, for the sake of comparing our training objectives and training strategy, both with respect to test set accuracy and risk certificates obtained.
    \item We connect PAC-Bayes with Backprop (PBB) methods to the Bayes-by-Backprop (BBB) method of \citet{blundell2015weight} which is inspired by Bayesian learning and achieved competitive test set accuracy. Unlike BBB, our training methods require less heuristics and also provide a risk certificate; not just an error estimate based on a test set.
\item We demonstrate via experimental results that PBB methods might be able to achieve self-certified learning with nontrivial certificates: obtaining competitive test set errors and computing non-vacuous bounds with much tighter values than previous works.
\end{enumerate}

\textbf{Broader Context.}
Deep learning is a vibrant research area.
The success of deep neural network models in several tasks has motivated many works that study their optimisation and generalisation properties, some of the collective knowledge is condensed in a few recent sources such as \citet{montavon2012neural,goodfellow2016book,aggarwal2018book}.
Some works focus on % 
experimenting with methods to train neural networks, others aim at generating knowledge and understanding about these fascinating learning systems. 
In this paper we intend to contribute both ways. 
We focus on % 
supervised classification problems through probabilistic neural networks, and we experiment with training objectives that are principled and consist of interpretable quantities.
Furthermore, our work puts an emphasis on certifying the quality of predictions beyond a specific data set.

Note that known neural network training methods range from those that have been developed based mainly on heuristics to those derived from sound principles.
Bayesian learning, for instance, offers principled approaches for learning data-dependent distributions over network weights (see e.g. \citealp{buntine1991bayesian}, \citealp{neal1992bayesian}, \citealp{mackay1992a_backpropagation}, \citealp{BarberBishop1997}), hence probabilistic neural nets arise naturally in this approach. 
Bayesian neural networks continue to be developed, with notable recent contributions e.g. by \citet{hernandez-lobato2015,martens_grosse2015,blundell2015weight,gal_ghahramani2016,louizos_welling2016,ritter2018,KhanLin2017,OsawaSKJETY2019,MaddoxIGVW2019}, among others. %
Our work is complementary of Bayesian learning in the sense that our methods also offer principled training objectives for learning probabilistic neural networks. However, there are differences between the PAC-Bayes and Bayesian learning approaches that are important to keep in mind (see our discussions in~\cref{s:PAC-Bayes} and~\cref{s:BBB}).
It is worth mentioning also that some works have pointed out the resemblance between PAC-Bayes bounds and the evidence lower bound (ELBO) of variational Bayesian inference (\citealp{AlquierRC2016,AchilleSoatto2018,thakur2019unifying,pitas2020}).
An insightful connection between Bayesian inference and the frequentist PAC-Bayes approach was discussed by \citet{germain-etal2016}.

As we pointed out before,
we are not the first to train a probabilistic neural network by minimising a PAC-Bayes bound, or to use a PAC-Bayes bound to give risk certificates for randomised neural nets.
We already mentioned \citet{LangfordCaruana2001} and 
\citet{dziugaite2017computing,dziugaite2018data}, whose works have directly influenced ours.\footnote{Note that \citet{LangfordCaruana2001} and 
\citet{dziugaite2017computing} called them \emph{stochastic} neural networks, arguably because the distribution over weights moves during training.
}
Next, we comment on some other works that connect PAC-Bayes with neural networks.
\citet{London2017} approached the generalisation of neural networks by a stability-based PAC-Bayes analysis, and proposed an adaptive sampling algorithm for SGD that optimises its distribution over training instances using multiplicative weight updates.  
\citet{neyshabur2017,neyshabur2018} examined the connection between some specifically defined complexity measures and generalisation, the part related to our work is that they specialised a form of the classic PAC-Bayes bound and used Gaussian noise on network weights to give generalisation bounds for probabilistic neural networks based on the norms of the weights.
\citet{zhou2019} compressed trained networks by pruning weights to a given target sparsity, and gave generalisation guarantees on the compressed networks,
which were based on randomising predictors according to their `description length' and a specialisation of a PAC-Bayes bound of \citet{catoni2007}.

We would like to point out that the present work builds on \cite{rivasplata2019pbb}. %
In the meantime, more works have appeared that connect neural networks with PAC-Bayes bounds in various settings: \citet{letarte2019dichotomize}, \citet{viallard2019interpreting}, \cite{lan2020}, possibly among others. 
We do not elaborate on these works as 
they deal with significantly different settings than ours.
The recent work by \citet{dziugaite2021role} is more closely related to ours in that they investigate the use of data to learn a PAC-Bayes prior. 

\smallskip

\textbf{Paper Layout.}
The rest of the paper is organised as follows. 
In Section~\ref{s:prelim} we briefly recall some notions of supervised learning, mainly to set the notation to be used later. 
In Section~\ref{s:PAC-Bayes} we outline the PAC-Bayes framework and discuss some PAC-Bayes bounds, while in Section~\ref{s:PBB} we present the training objectives derived from them. Section~\ref{s:BBB} discusses the connection between our work and \citet{blundell2015weight}. 
The technical \cref{s:kl_invert} describes the binary KL inversion strategy and the ways we use it.
In Section~\ref{s:exp} we present our experimental results. 
We conclude and discuss future research directions in Section~\ref{s:conclude}.

\section{Generalisation through Risk Upper Bounds}
\label{s:prelim}

An algorithm that trains a neural network receives a finite list of training examples and produces a data-dependent weight vector $\hat{w} \in \cW \subset \mathbb{R}^p$, which is used to make predictions on unseen examples.
The ultimate goal is for the algorithm to find a weight vector that generalises%
\footnote{In statistical learning theory the meaning of \emph{generalisation} of a learning method has a precise definition (see e.g. \citealp{SSBD2014}). We use the word in a slightly broader sense here.
}
 well, meaning that the decisions arrived at by using the learned $\hat{w}$ should give rise to a small loss on unseen examples
from the same distribution as the training data.
Turning this into precise statements requires a formal description of the learning setting, briefly discussed next. The  reader familiar with learning theory can skip the next couple of paragraphs and come back if they need clarifications regarding notation.

The training algorithm receives a size-$n$ random sample $S = (Z_1,\ldots,Z_n)$. Each example $Z_i$ is randomly drawn from a space $\cZ$ according to an underlying (but unknown) probability distribution%
\footnote{$\cM_1(\cZ)$ denotes the set of all probability measures over $\cZ$.}
$P \in \cM_1(\cZ)$. 
The example space usually takes the form $\cZ = \cX\times\cY$ in supervised learning, where $\cX\subset\R^d$ and $\cY\subset\R$, each example being a pair $Z_i = (X_i,Y_i)$ consisting of an input $X_i$ and its corresponding label $Y_i$. 
A space $\cW \subseteq \R^p$ encompasses all possible weights, and
it is understood that each possible weight vector $w \in \cW$ maps to a predictor function $h_w: \cX\to\cY$ that will assign a label $h_{w}(X) \in \cY$ to each new input  $X \in\cX$. 
While statistical inference is largely concerned with elucidating properties of the unknown data-generating distribution, the main focus of machine learning is on the quality of predictions, measured by the \emph{expected loss} on unseen examples, also called the \emph{risk}:
\begin{align}
\label{eq:risk}
    L(w) = \EE[\ell(w,Z)] = \int_{\cZ} \ell(w,z) P(dz)    \,.
\end{align}
Here $\ell : \cW\times\cZ\to[0,\infty)$ is a fixed loss function. 
With these components, regression is defined as the problem when $\cY = \R$ and the loss function is the squared loss, namely $\ell(w,z) = (y-h_w(x))^2$ where $z = (x,y)$ is the input-label pair,
while binary classification is the problem where $\cY = \{0,1\}$ (or $\cY = \{-1,+1\}$) and the loss is set to be the zero-one loss:  $\ell(w,z) = \mathbb{I}[y \neq h_w(x)]$.

The goal of learning is to find a weight vector with small risk $L(w)$.
However, since the data-generating distribution $P$ is unknown, $L(w)$ is an unobservable objective. 
Replacing the expected loss with the average loss on the data gives rise 
to an observable objective called the \emph{empirical risk} functional:
\begin{align}
\label{eq:empirical_risk}
    \hL_S(w)
    = \frac{1}{n}\sum_{i=1}^{n} \ell(w,Z_i)    \,.
\end{align}
In practice, the minimisation of $\hL_S$ is often done with some version of gradient descent. Since the zero-one loss gives rise to a piecewise constant loss function, which is provably hard to optimise, in classification it is common to replace it with a smooth(er) loss, such as the cross-entropy loss, while changing the range of $h_w$ to $[0,1]$.

Under certain conditions, a small empirical risk leads to a weight that is guaranteed to have a small risk gap\footnote{The risk gap is the difference between the risk \eqref{eq:risk} and the empirical risk \eqref{eq:empirical_risk}. 
}. 
Examples of such conditions are when the set of functions $\{h_w \,:\, w\in \R^p\}$ representable has a small capacity relative to the sample size, or the map that produces the weights given the data is stable.
However, often minimising the empirical risk can lead to a situation where the risk of the learned weight is significantly larger than the empirical risk ---a case of overfitting.
To prevent overfitting, various methods are commonly used. These include complexity regularisation, early stopping, injecting noise in various places into the learning process, among others (e.g. \citealp{srivastava2014dropout}, \citealp{wan2013regularization}, \citealp{CaruanaLG2000}, \citealp{Hinton1993_colt,Hinton1993_icann}).

An alternative to these is to minimise a surrogate objective which is guaranteed to give an upper bound on the risk. As long as the upper bound is tight and the optimisation gives rise to a small value for the surrogate objective, the user can be sure that the risk will also be small: In this sense, overfitting is automatically prevented, while we also automatically get a self-bounding learning method (cf. \citealp{Freund1998,langford2003microchoice}). 
In this paper we follow this last approach, with two specific training objectives derived from corresponding PAC-Bayes bounds, which we introduce in the next section. 
The approach to learning data-dependent distributions over hypotheses by minimising a PAC-Bayes bound was mentioned already by \citet{McAllester1999}, credit for this approach in various contexts is due also to \citet{germain-etal2009}, \citet{SeldinTishby2010}, \citet{Keshet-etal2011}, \citet{NoyCrammer2014robust}, \citet{Keshet-etal2017},
among others. 
\citealp{dziugaite2017computing} used this approach for training probability distributions over neural network weights, and used the PAC-Bayes-kl bound for computing numerical risk bound values for the corresponding randomised classifiers. 
The work of Dziugaite and Roy brought to attention that their strategy delivers non-vacuous risk bound values for randomised neural network classifiers in the regime where the models have (many) more parameters than training data.

As will be demonstrated below, our experiments based on our two training objectives 
$f_{\mathrm{quad}}$ and $f_{\mathrm{lambda}}$ (Eq.~\eqref{eq:obj-quad} and Eq.~\eqref{eq:obj-lambda} below) lead to (a) test set performance comparable to that of the Bayesian learning method used by \citet{blundell2015weight}, while (b) computing non-vacuous bounds with tighter values than those obtained by $f_{\mathrm{classic}}$ (Eq.~\eqref{eq:obj-classic} below) which is essentially equivalent to the training objective used by Dziugaite and Roy.

\section{PAC-Bayes Bounds}
\label{s:PAC-Bayes}

Probabilistic neural networks are realised as probability distributions over the weight space. 
While the outcome of a classical (non-probabilistic) neural network training method is a fixed (but data-dependent) setting of the weights, the outcome of training a probabilistic neural network is a data-dependent distribution\footnote{Formally, a data-dependent distribution over $\cW$ is a stochastic kernel from $\cS$ to $\cW$. This formalisation of data-dependent distributions over predictors was covered recently by \citet{rivasplata2020beyond}.}
$Q_S$ over network weights.

For a given distribution $Q$ over network weights, the randomised classifier is defined as follows:
Given a fresh input $X$, the randomised classifier predicts its label by drawing a weight vector $W$ at random according to $Q$ and applying the predictor $h_W$ to $X$.
Each new prediction requires a fresh draw.
The randomised predictor is identified with the distribution $Q$ that defines it, for simplicity of notation.

One way, which we adopt in this paper, to measure the performance of the randomised predictor corresponding to the distribution $Q$ over $\cW$ is to use the $Q$-weighted losses, since these are the expected losses over the random draws of weights defining the randomised predictor. 
Accordingly, the population loss of $Q$ becomes
\begin{align}
\label{eq:population_loss_Q}
L(Q) = \int_{\cW} L(w) Q(dw)    \,,
\end{align}
and the empirical loss of $Q$ becomes
\begin{align}
\label{eq:empirical_loss_Q}
\hL_S(Q) = \int_{\cW} \hL_S(w) Q(dw)  \,.
\end{align}
These definitions extend the loss notions $L(w)$ and $\hL_S(w)$ previously defined for a given weight $w$, to corresponding notions $L(Q)$ and $\hL_S(Q)$ for a given distribution $Q$ over weights.
Then, PAC-Bayes bounds relate the population loss
$L(Q)$ to the empirical loss $\hL_S(Q)$ and other quantities, by means of inequalities that hold with high probability.

\smallskip

To introduce the promised PAC-Bayes bounds we need to recall some further definitions.
Given two probability distributions $Q, Q^{\prime} \in \cM_1(\cW)$, the Kullback-Leibler (KL) divergence of $Q$ from $Q^{\prime}$, also known as relative entropy of $Q$ given $Q^{\prime}$, is defined as follows:
\begin{align*}
% \vspace{-4mm}
\KL(Q \Vert Q^{\prime}) 
= \int_{\cW} \log\Bigl( \frac{dQ}{dQ^{\prime}} \Bigr) \,dQ  \,.
% \vspace{-4mm}
\end{align*} 
% when $dQ/dQ^{\prime}$, the Radon-Nikodym derivative of $Q$ with respect to $Q^{\prime}$, is defined; otherwise  $\KL(Q \Vert Q^{\prime})=\infty$.  
This equation defines $\KL(Q \Vert Q^{\prime})$ when $dQ/dQ^{\prime}$, the Radon-Nikodym derivative of $Q$ with respect to $Q^{\prime}$, is defined; otherwise  $\KL(Q \Vert Q^{\prime})=\infty$.  
For $q,q^\prime\in [0,1]$ we define
\begin{align}\label{eq:binkl}
    \kl(q \Vert q^\prime) = q \log(\frac{q}{q^\prime}) + (1-q)\log(\frac{1-q}{1-q^\prime})   \,,
\end{align}
which is  called the binary KL divergence, and is the divergence of the Bernoulli distribution with parameter $q$ from the Bernoulli distribution with parameter $q^\prime$.

\smallskip

The PAC-Bayes-kl inequality, originally called the PAC-Bayes relative entropy bound \citep{LangfordSeeger2001,seeger2002} is a fundamental result from which some other PAC-Bayes bounds may be derived. We state this result next for easy reference.

\begin{theorem}[PAC-Bayes-kl]
\label{thm:pb-kl}
Let the triple $(\cW,\cZ,\ell)$ consist of a weight space $\cW \subset \R^p$, an example space $\cZ$, and a loss function $\ell : \cW\times\cZ \to [0,1]$.
Let $n$ be a positive integer, and 
let $\hL : \cZ^n\times\cW \to [0,1]$ be the empirical risk functional defined as $\hL(s,w) = n^{-1} \sum_{i=1}^n \ell(w,z_i)$ for $s = (z_1,\ldots,z_n) \in \cZ^n$; and write $\hL_s(w) = \hL(s,w)$.
Let $P \in \cM_1(\cZ)$, and let the risk $L : \cW \to [0,1]$ be the functional defined as $L(w) = \E[\ell(w,Z)]$ with $Z \sim P$.

Then,
for any data-free distribution $Q^0$ over $\cW$,
and for any $\delta \in (0,1)$,
with probability of at least $1-\delta$ over size-$n$ i.i.d. samples $S \sim P^{\otimes n}$, 
simultaneously for all distributions $Q$ over $\cW$ we have
\begin{align}
    \kl(\hL_S(Q) \Vert L(Q)) \leq
    \frac{\KL(Q \Vert Q^0)+\log(\frac{ 2\sqrt{n} }{\delta})}{n}\,.
\label{eq:pb-kl}
\end{align}
\end{theorem}
The assumption that $Q^0$ is a data-free distribution over $\cW$  means that $Q^0$ is fixed without any dependence on the data on which the bound is evaluated (to be very specific, $Q^0$ cannot depend on the sample $S$ on which $\hL_S$ is evaluated).
The original form of this bound that was derived by \cite{LangfordSeeger2001} has a slightly different dependence on $n$, the form presented here has the sharp dependence on $n$ as clarified by \cite{maurer2004note}.

\smallskip

The PAC-Bayes-kl bound can be relaxed in various ways to obtain other PAC-Bayes bounds [see e.g. \citealp{TolstikhinSeldin2013}].
For instance, using the well-known version of Pinsker's inequality $\kl(\hat{p}\Vert p) \ge 2(p-\hat p)^2$ one can lower-bound the binary KL divergence, and then solve the resulting inequality for $L(Q)$, which leads to a PAC-Bayes bound of equivalent form to that of the classic bound of \cite{McAllester1999}, hence we shall call it the 
\textbf{PAC-Bayes-classic} bound:
for any $\delta \in (0,1)$,
with probability of at least $1-\delta$ over size-$n$ i.i.d. random samples $S$,
simultaneously for all distributions $Q$ over $\cW$ we have
\begin{align}
L(Q) \leq \hL_S(Q) +
\sqrt{\frac{\KL(Q \Vert Q^0)+\log(\frac{ 2\sqrt{n} }{\delta})}{2n}} \,.
\label{eq:pb-classic}
\end{align}
Notice that the PAC-Bayes-classic bound is an inequality that holds simultaneously for all distributions $Q$ over weights, with high probability (over samples). 
In particular, the upper bound may be optimised to choose a distribution $Q$ in a data-dependent manner.
At a high level, the interest in finding a $Q$ for which the upper bound is minimal is because then this may guarantee a small $L(Q)$, since \cref{eq:pb-classic} gives an upper bound on $L(Q)$.

\smallskip

An alternative way to relax the PAC-Bayes-kl bound is using
the refined version of Pinsker's inequality $\kl(\hat{p} \Vert p) \geq (p - \hat{p})^2/(2p)$ valid for $\hat{p}<p$ 
[see e.g. \citealp[Lemma 8.4]{boucheron2013concentration}], 
which is tighter than the former version when $p < 1/4$,
and this refined inequality gives
\begin{align}
    L(Q) \leq 
    \hL_S(Q) + 
    \sqrt{ 
    2L(Q)\frac{\KL(Q \Vert Q^0)+\log(\frac{2\sqrt{n}}{\delta})}{n}
    }\,. \tag{$\star$}
\label{eq:star}
\end{align}
The difference to the result one gets from the well-known version of Pinsker's inequality is the appearance of $L(Q)$ under the square root on the right hand side. This, in particular, tells us that the inequality is tighter than \cref{eq:pb-classic} when the population loss, $L(Q)$, is smaller (specifically when $L(Q) < 1/4$). But it is exactly because of the appearance of $L(Q)$ on the right-hand side that this bound is not immediately useful for optimisation purposes.
However, one can view the inequality \eqref{eq:star} as a quadratic inequality on $\sqrt{L(Q)}$.
Solving this inequality for $L(Q)$ leads to our  \textbf{PAC-Bayes-quadratic} bound which to the best of our knowledge is new \citep{rivasplata2019pbb}:
for any $\delta \in (0,1)$,
with probability of at least $1-\delta$ over size-$n$ i.i.d. random samples $S$,
simultaneously for all distributions $Q$ over $\cW$ we have
\begin{align}
\label{eq:pb-quad}
    L(Q) 
    &\leq \left(
    \sqrt{ 
    \hL_S(Q) + \frac{\KL(Q \Vert Q^0) + \log(\frac{2\sqrt{n}}{\delta})}{2n} 
    } \right.
    +
    \left.
    \sqrt{ 
    \frac{\KL(Q \Vert Q^0) + \log(\frac{2\sqrt{n}}{\delta})}{2n} 
    } \right)^2\,.
\end{align}
Similarly to the PAC-Bayes-classic bound, the PAC-Bayes-quadratic bound is an inequality that holds uniformly over all $Q$; hence the upper bound may be optimised with respect to $Q$ in order to obtain a data-dependent distribution over weights. 

Alternatively, 
using \eqref{eq:star} combined with the inequality $\sqrt{ab} \leq \tfrac{1}{2}(\lambda a + \frac{b}{\lambda})$ valid for all $\lambda > 0$, after some derivations one obtains the %so-called 
\textbf{PAC-Bayes-$\lambda$} bound of \citet{thiemann-etal2017}: 
for any $\delta \in (0,1)$,
with probability of at least $1-\delta$ over size-$n$ i.i.d. random samples $S$, 
simultaneously for all distributions $Q$ over $\cW$ and $\lambda\in(0,2)$ we have
\begin{align}
L(Q)
\le 
\frac{\hL_S(Q)}{1-\lambda/2}+ \frac{\KL(Q \Vert Q^0)+\log(2\sqrt{n}/\delta)}{n\lambda(1-\lambda/2)}\,.
\label{eq:pb-lambda}
\end{align}
An interesting feature of the PAC-Bayes-$\lambda$ bound is that
this bound holds uniformly over all $Q$ and $\lambda \in (0,2)$, hence in principle this bound is optimisable over both these quantities.
The work of \citet{thiemann-etal2017} discussed conditions under which the PAC-Bayes-$\lambda$ bound can be optimised alternatingly over $Q$ and $\lambda$.
Since this bound holds uniformly over $\lambda \in (0,2)$, it is possible to search a grid of $\lambda$-values without worsening the bound.

\smallskip

Each of the previous three PAC-Bayes bounds (\cref{eq:pb-classic}, \cref{eq:pb-quad} and \cref{eq:pb-lambda}) is an upper bound on $L(Q)$ that holds simultaneously for all distributions $Q$ over weights, with high probability (over samples). 
In particular, the bounds allow to choose a distribution $Q_S$ in a data-dependent manner, which is why they are usually called `posterior' distributions in the PAC-Bayesian literature.
However, these distributions should not be confused with Bayesian posteriors.
Note that in the frequentist PAC-Bayes learning approach, what is called `prior' is a reference distribution, and what is called `posterior' is an unrestricted distribution, in the sense that there is no likelihood-type factor connecting these two distributions.
Having said that, \citet{germain-etal2016} showed that the optimal PAC-Bayes posterior coincides with the Bayesian posterior when the loss function is the negative log-likelihood.
However, in general, when learning `posteriors' via PAC-Bayes with Backprop, there is no place for asking whether ``our posterior approximates the true posterior'' on account that the notion of true PAC-Bayes posterior is nonexistent.

\smallskip

Below in Section~\ref{s:PBB} we discuss  training objectives derived from these bounds. Notice that there are many other PAC-Bayes bounds available in the literature; the usual ones are by \citet{McAllester1999}, \citet{LangfordSeeger2001}, \citet{catoni2007}; but see also \citet{McAllester2003}, \citet{Keshet-etal2011}, and \cite{McAllester2013}.
Each bound readily leads to a training objective by replicating our procedure (described in \cref{s:PBB} below).
Some references for several kinds of PAC-Bayes bounds are \citet{germain-etal2009} and \cite{begin2014pac,begin2016pac}, the mini tutorial of \citet{vanerven2014mini}, and the primer of \citet{guedj2019primer}.

\section{The Bayes by Backprop (BBB) Objective}
\label{s:BBB}

The `Bayes by backprop' (BBB) method % 
of \citet{blundell2015weight} is inspired by a variational Bayes argument \citep{JordanGJS98,fox2012tutorial}, where
the idea is to learn a distribution over weights that approximates the Bayesian posterior distribution. Choosing a $p$-dimensional Gaussian $Q_{\theta}= \mu + \sigma \mathcal{N}(0,I)$, parametrised by $\theta=(\mu,\sigma) \in \R^p \times \R^p$, the optimum parameters are those that minimise  $\KL(Q_{\theta} \Vert P(\cdot|S))$, i.e. the KL divergence from $Q_{\theta}$ and the Bayesian posterior $P(\cdot|S)$. 
By a simple calculation, and using the Bayes rule, one can extract:
\begin{align*}
    \KL(Q_{\theta} \Vert P(\cdot|S)) 
    =  \int_{\cW} -\log( P(S|w)) Q_{\theta}(dw)
        + \KL(Q_{\theta} \Vert Q^0) \,,
\end{align*}
where $Q^0$ stands here for the Bayesian prior distribution.
Thus, minimising $\KL(Q_{\theta} \Vert P(\cdot|S))$ is equivalent to minimising the
 right-hand side, which presents a sum of a data-dependent term (the expected negative log-likelihood) and a prior-dependent term ($\KL(Q_{\theta} \Vert Q^0)$). This optimisation problem is analogous to that of minimising a PAC-Bayes bound, since the latter balances a fit-to-data term (the empirical loss) and a fit-to-prior term (the KL).

 There is indeed a close connection between the PAC-Bayes and Bayesian learning approaches, as has been pointed out by the work of \cite{germain-etal2016}, when the loss function is the negative log-likelihood.
 Beyond this special case, the PAC-Bayes learning approach offers more flexibility in design choices, such as the choice of loss functions and the choice of distributions.
 This is because the PAC-Bayes `prior' is a reference distribution and the PAC-Bayes `posterior' does not need to be derived from a prior by a likelihood update factor.
 This is a crucial difference with Bayesian learning, and one that makes the PAC-Bayes framework a lot more flexible in the choice of distributions over parameters, even compared to generalised Bayesian approaches \citep{bissiri2016general}.

As we mentioned before, the training objective proposed by \citet{blundell2015weight} is inspired by the variational Bayesian argument outlined above, in particular, in our notation the training objective they proposed and experimented with is as follows:
\begin{align}
f_{\mathrm{bbb}}(Q) = \hL_S(Q) + \eta\, \frac{ \KL( Q \Vert Q^0 ) }{n}\,.
\label{eq:bbbobjective}
\end{align}
The scaling factor,
$\eta>0$,  is introduced in a heuristic manner to make the method more flexible, 
 while the variational Bayes argument gives \eqref{eq:bbbobjective} with $\eta=1$. 
When $\eta$ is treated as a tuning parameter, the method can be interpreted as searching in ``KL balls'' 
centered at $Q^0$ of various radii. Thus, the KL term then plays the role of penalising the complexity of the model space searched.
\citet{blundell2015weight} proposed to optimise this objective (for a fixed $\eta$) 
using stochastic gradient descent (SGD), which randomises over both mini-batches 
and the weights, and used the pathwise gradient estimate \citep{price1958useful}. 
The resulting gradient-calculation procedure can be seen to be only at most twice as expensive as standard backpropagation ---hence the name of their method. 
The hyperparameter $\eta>0$ is chosen using a validation set, which is also often used to select the best performing model among those that were produced during the course of running SGD (as opposed to using the model obtained when the optimisation procedure finishes).

\section{Towards Practical PAC-Bayes with Backprop (PBB) Methods}
\label{s:PBB}

The essential idea of `PAC-Bayes with Backprop' (PBB) is
to train a probabilistic neural network by minimising a PAC-Bayes bound via stochastic gradient descent (SGD) optimisation. 
Here we present two training objectives, derived from Eq.~\eqref{eq:pb-quad} and Eq.~\eqref{eq:pb-lambda} respectively, in the context of \emph{classification problems} when the loss is the zero-one loss or a surrogate loss. These objectives are used here for the first time to train probabilistic neural networks. We also discuss the training objective derived from Eq.~\eqref{eq:pb-classic} for comparison purposes.

To optimise the weights of neural networks the standard idea is to use a form of stochastic gradient descent,
which requires the ability to efficiently calculate gradients of the objective to be optimised. 
When the loss is the zero-one loss, 
the training loss viewed as a function of the weights, $w \mapsto \hL^{01}_S(w)$, is piecewise constant, which makes simple gradient-based methods fail (since the gradient, whenever it exists, is zero). 
As such, it is customary to replace the zero-one loss with a smoother ``surrogate loss'' that plays well with gradient-based optimisation.
In particular, the standard loss used on multiclass classification problems is the cross-entropy loss, $\ell^{\ce}:\R^k \times [k] \to \R$ defined by $\ell^{\ce}(u,y) = -\log(\sigma(u)_y)$  where $u\in\R^k$, $y\in [k] = \{1,\dots,k\}$ and $\sigma:\R^k \to [0,1]^k$ is the soft-max function defined by $\sigma(u)_i = \exp(u_i)/\sum_{j}\exp(u_j)$.
This choice can be justified on the grounds that $\ell^{\ce}(u,y)$ gives an upper bound on the probability of mistake when the label is chosen at random from the distribution produced by applying soft-max on $u$ (e.g., $u=$ the output of the last linear layer of a neural network).%
\footnote{Indeed, 
owning to the inequality $\log(x)\le x-1$, which is valid for any $x>0$, 
given any $u\in \R^k$ and $y\in [k]$, if
$Y\sim \sigma(u)$ then 
$\EE[\mathbb{I}\{Y\ne y\}] = \mathbb{P}(Y\ne y) = 1-\sigma(u)_y \le \ell^{\ce}(u,y)$.
}
We thus also propose to replace the zero-one loss with the cross-entropy loss in either \cref{eq:pb-quad} or \cref{eq:pb-lambda}, leading to the objectives
\begin{align}
\label{eq:obj-quad}    
    &\hspace*{-3mm} f_{\mathrm{quad}}(Q) = 
    \left(
    \sqrt{ 
    \hL^{\ce}_S(Q) + \frac{\KL(Q \Vert Q^0) + \log(\frac{2\sqrt{n}}{\delta})}{2n} 
    } \right. 
    +
    \left.
    \sqrt{ 
    \frac{\KL(Q \Vert Q^0) + \log(\frac{2\sqrt{n}}{\delta})}{2n} 
    } \right)^2
\end{align}
and
\begin{align}
\label{eq:obj-lambda}
    &\hspace*{-3mm} f_{\mathrm{lambda}}(Q,\lambda) =
    \frac{\hL^{\ce}_S(Q)}{1-\lambda/2}+ \frac{\KL(Q \Vert Q^0)+\log(2\sqrt{n}/\delta)}{n\lambda(1-\lambda/2)}~.
\end{align}
For comparison, the training objective derived from \cref{eq:pb-classic} takes the following form:
\begin{align}
\label{eq:obj-classic}
    &\hspace*{-3mm} f_{\mathrm{classic}}(Q) =
    \hL^{\ce}_S(Q) + 
    \sqrt{\frac{\KL(Q \Vert Q^0) + \log(\frac{2\sqrt{n}}{\delta}) }{2n}}~.
\end{align}
Here, $\hL^{\ce}_S(w) = \frac1n \sum_{i=1}^n \tilde{\ell}^{\ce}_1(h_w(X_i),Y_i)$ %
is the empirical error rate under the `bounded' version of cross-entropy loss, namely the loss $\tilde{\ell}^{\ce}_1$ described next, and $h_w:\cX \to \R^k$ denotes the function implemented by the neural network that uses weights $w$.
The next issue to address is that the cross-entropy loss is unbounded, while the PAC-Bayes bounds that inspired these objectives require a bounded loss with range $[0,1]$. 
This is fixed by enforcing an upper bound 
on the cross-entropy loss by lower-bounding the network probabilities by a value $p_\mathrm{min}>0$ \citep{dziugaite2018data}.
This is achieved by replacing $\sigma$ in the definition of $\ell^{\ce}$ by $\tilde \sigma(u)_y = \max( \sigma(u)_y, p_{\min})$.
This adjustment gives a `bounded cross-entropy' loss function
$\tilde{\ell}^{\ce}(u,y) = -\log(\tilde{\sigma}(u)_y)$
with range between 0 and $\log(1/p_{\min})$.
Finally, re-scaling by $1/\log(1/p_{\min})$ gives a loss function $\tilde{\ell}^{\ce}_1$ with range [0,1] ready to be used in the PAC-Bayes bounds and training objectives discussed here. 
The latter ($\tilde{\ell}^{\ce}_1$) is used as the surrogate loss for training in all our experiments with $f_{\mathrm{quad}}$, $f_{\mathrm{lambda}}$, and $f_{\mathrm{classic}}$.

\subsection{Optimisation Problem}
Optimisation of $f_{\mathrm{quad}}$ and $f_{\mathrm{classic}}$ (Eq.~\eqref{eq:obj-quad} and Eq.~\eqref{eq:obj-classic}) entails minimising over $Q$ only, while
optimisation of $f_{\mathrm{lambda}}$ (Eq.~\eqref{eq:obj-lambda}) is done by alternating minimisation with respect to  $Q$ and $\lambda$, similar to the procedure that was used by \citet{thiemann-etal2017} in their experiments with SVMs. 
By choosing $Q$ appropriately, in either case we use the pathwise gradient estimator 
\citep{price1958useful,jankowiak2018pathwise,JMLR:v21:19-346}
as done by \citet{blundell2015weight}.
In particular, assuming that 
 $Q=Q_{\theta}$ with $\theta\in \R^q$ is such that 
$h_W(\cdot)$ with $W \sim Q_{\theta}$ $(W\in \R^p$) 
has the same distribution as $h_{f_\theta(V)}(\cdot)$ where $V\in \R^{p^\prime}$ is drawn at random from a \emph{fixed} distribution $P_V$ and $f_\theta: \R^{p^\prime} \to \R^p$ is a smooth map, an unbiased estimate of the gradient of the loss-map
$\theta \mapsto Q_{\theta} [ \ell( h_{\bullet}(x), y ) ]$ at some $\theta$ can be obtained by drawing $V\sim P_V$ and calculating $ \frac{\partial}{\partial \theta}  \ell( h_{f_\theta(V) }(x), y) $, thereby reducing the efficient computation of the gradient to the application of the backpropagation algorithm on the map $\theta \mapsto  \ell( h_{f_\theta(v) }(x), y)$ at $v=V$.%
\footnote{Indeed,
 $\frac{\partial}{\partial \theta} \int Q_{\theta}(dw) \ell( h_{w}(x), y ) 
= \frac{\partial}{\partial \theta} \int P_V(dv) \ell( h_{f_\theta(v)}(x), y ) 
= \int P_V(dv) \frac{\partial}{\partial \theta}  \ell( h_{f_\theta(v) }(x), y )$, where the interchange of the partial derivative and the integral is justified when the partial derivatives are integrable, which needs to be verified on a case-by-case basis.
See e.g. \citet{ruiz2016}.
}

In our experiments the PAC-Bayes posterior is parametrised as a diagonal Gaussian distribution over weight space $\cW = \R^p$. Then a sample of the posterior can be obtained by sampling a standard Gaussian, scaling each coordinate by a corresponding standard deviation from the vector $\sigma = (\sigma_i)_{i \in [p]} \in \R^p$, and shifting by a mean vector $\mu \in \R^p$.
We parametrise $\sigma$ coordinatewise as $\sigma = \log(1+\exp(\rho))$ so $\sigma$ is always non-negative. 
Following \citet{blundell2015weight}, the reparametrisation we use is $W = \mu + \sigma \odot V$ with appropriate distribution (Gauss or Laplace) for each coordinate of $V$, although other reparametrisations are possible \citep{OsawaSKJETY2019,KhanLin2017}. %
Gradient updates are with respect to vectors $\mu$ and $\rho$, 
as can be seen in Algorithm~\ref{PBWBP}. Note that after sampling the weights,  the gradients for the mean and standard deviation are shared and are exactly the gradients found by the usual backpropagation algorithm on a neural network. More specifically, to learn both the mean and the standard deviation we simply calculate the usual gradients found by backpropagation, and then scale and shift them as done by \cite{blundell2015weight}.
Note that \cref{PBWBP} shows the procedure for optimising $f_{\mathrm{quad}}$ with Gaussian noise. The procedure with Laplace noise is similar. The procedure for $f_{\mathrm{classic}}$ is similar. The procedure for $f_{\mathrm{lambda}}$ would be very similar except that $f_{\mathrm{lambda}}$ has the additional parameter $\lambda$.

% \todoo[inline]{Insert here comment/discussion about when f-quad is tighter than f-classic. [The discussion about the two Pinsker inequalities rightfully belongs in Section 3 tho] ... One can compare these two inequalities, to find regime of $p,\hat{p}$ in which one is better than the other. The result of the comparison is that Eq.~\eqref{eq:pinsker} (used in $f_{\mathrm{classic}}$) is tighter whenever $p > 1/4$, and Eq.~\eqref{eq:pinsker_refined} (used in $f_{\mathrm{quad}}$) is tighter whenever $p < 1/4$.
% They match if $p = 1/4$.}

\smallskip

As discussed in \cref{s:PAC-Bayes}, the PAC-Bayes bounds from which these training objectives 
% ($f_{\mathrm{classic}}$, $f_{\mathrm{quad}}$, and $f_{\mathrm{lambda}}$) 
were derived are relaxations of the PAC-Bayes-kl bound (\cref{thm:pb-kl}).
We refer the reader to Eq.~\eqref{eq:binkl} for the definition of the binary KL divergence, denoted $\kl(\cdot\Vert\cdot)$.
It was explained that $f_{\mathrm{classic}}$ is a relaxation of PAC-Bayes-kl bound obtained by Pinsker's inequality: 
\begin{equation}
    \kl(\hat{p}\Vert p) \ge 2(p-\hat p)^2
    \hspace{5mm} \text{for} \hspace{1mm} \hat{p}, p \in (0,1)\,. \hspace{12mm}
\label{eq:pinsker}
\end{equation}
On the other hand, $f_{\mathrm{quad}}$ and $f_{\mathrm{lambda}}$ are relaxations of the PAC-Bayes-kl bound obtained using the refined version Pinsker's inequality:
\begin{equation}
    \kl(\hat{p} \Vert p) \geq \frac{(p - \hat{p})^2}{2p}
    \hspace{5mm} \text{for} \hspace{1mm} \hat{p}, p \in (0,1), \hspace{1mm} \hat{p}<p\,.
\label{eq:pinsker_refined}
\end{equation}
One can compare these two inequalities, to find regime of $p,\hat{p}$ in which one is better than the other. The result of the comparison is that Eq.~\eqref{eq:pinsker} 
% (used in $f_{\mathrm{classic}}$) 
is tighter whenever $p > 1/4$, and Eq.~\eqref{eq:pinsker_refined} 
% (used in $f_{\mathrm{quad}}$) 
is tighter whenever $p < 1/4$.
They match if $p = 1/4$.
This comparison might be relevant for understanding the differences---in terms of tightness of risk certificates but also test performance---between the solutions found by these training objectives.

\begin{algorithm}[t]
\scriptsize
 \caption{PAC-Bayes with Backprop (PBB)} 
 \label{PBWBP}
    \begin{algorithmic}[1] %
     \Require
     \Statex $\mu_0$ \Comment{Prior center parameters} %
     \Statex $\rho_0$ \Comment{Prior scale hyper-parameter}
     \Statex $Z_{1:n}$ \Comment{Training examples (inputs + labels)}
     \Statex $\delta \in (0,1)$
              \Comment{Confidence parameter}      
     \Statex $\alpha \in (0,1)$, \ $T$ \Comment{Learning rate; Number of iterations}
     \Ensure {Optimal $\mu \in \mathbb{R}^p, \; \rho \in \mathbb{R}^p$  } \Comment{Posterior centers and scales} %
        \Procedure{pb\_quad\_gauss}{} %
     \State $\mu \gets  \mu_0$ \Comment{Set initial posterior center to prior center}
     \State $\rho \gets  \rho_0$ \Comment{Set initial posterior scale to prior scale}
            \For{$t\gets 1:T$} \Comment{Run SGD for T iterations.}
                 \State Sample $ V \sim \mathcal{N}(0,I)$
                 \State $W = \mu + \log(1+\exp(\rho)) \odot V$
                 \State $f = f_{\mathrm{quad}}(Z_{1:n}, W, \mu, \rho, \mu_0, \rho_0, \delta) $
                 \vspace{1mm} 
                    \State SGD gradient step using $\begin{bmatrix} \nabla_\mu f \\ \nabla_\rho f  \end{bmatrix}$, \;\;\;
{                     $\nabla_\mu f = \frac{\partial f}{\partial W} + \frac{\partial f}{\partial \mu}$, \;\;\;
                    $\nabla_\rho f = \frac{\partial f}{\partial W} \cdot \frac{V}{1+\exp(-\rho)} + \frac{\partial f}{\partial \rho}$}

            \EndFor 
            \State \textbf{return} $\mu, \rho$
        \EndProcedure
    \end{algorithmic}
\end{algorithm}

\subsection{The Choice of the PAC-Bayes Prior Distribution}

We experiment both with priors centered at randomly initialised weights and priors learnt by empirical risk minimisation using the surrogate loss on a subset of the data set which is independent of the subset used to compute the risk certificate. Note
that all $n$ training data are used by the learning algorithm ($n_0$ examples used to build the prior, $n$ to learn the posterior and $n - n_0$ to evaluate the risk certificate). This is to avoid needing differentially private arguments to justify learning the prior \citep{dziugaite2018data}. Since the posterior is initialised to the prior, the learnt prior translates to the posterior being initialised to a large region centered at the empirical risk minimiser. Similar approaches for building data-dependent priors have been considered before in the PAC-Bayesian literature \citep{ambroladze2007tighter, JMLR:v13:parrado12a}.

For our PAC-Bayes prior over weights we experiment with Gaussian and with Laplace distributions. In each case, the PAC-Bayes posterior learnt by PBB is of the same kind (Gaussian or Laplace) as the prior.
Next we give formulas for computing the KL term in our training objectives for each of these distributions.

\subsubsection{Formulas for the KL: Laplace and Gaussian}
\label{s:L_vs_G}

The Laplace density with mean parameter $\mu \in \R$ and with variance $b>0$ is the following: 
\begin{align*}
    p(x) = (2b)^{-1} \exp\bigl(-\frac{|x-\mu|}{b}\bigr)\,.
\end{align*}
The KL divergence for two Laplace distributions is 
% as follows:
\begin{align}
\label{eq:KL_Laplace}
    \KL(\operatorname{Lap}(\mu_1,b_1) \Vert \operatorname{Lap}(\mu_0,b_0))
    = \log(\frac{b_0}{b_1}) + \frac{|\mu_1-\mu_0|}{b_0} + \frac{b_1}{b_0}e^{-|\mu_1-\mu_0|/b_1} - 1\,.
\end{align}

For comparison, recall that the Gaussian density with mean parameter $\mu \in \R$ and variance $b>0$ has the following form:
\begin{align*}
   p(x) = (2\pi b)^{-1/2} \exp\bigl(-\frac{(x-\mu)^2}{2b}\bigr)\,.
\end{align*}
The KL divergence for two Gaussian distributions is 
% as follows:
\begin{align}
\label{eq:KL_Gauss}
    &\KL(\operatorname{Gauss}(\mu_1,b_1) \Vert \operatorname{Gauss}(\mu_0,b_0))
    = \frac{1}{2}\Bigl(\log(\frac{b_0}{b_1}) + \frac{(\mu_1-\mu_0)^2}{b_0} + \frac{b_1}{b_0}   - 1 \Bigr)\,.
\end{align}

The formulas \eqref{eq:KL_Laplace} and \eqref{eq:KL_Gauss} above are for the KL divergence between one-dimensional Laplace or Gaussian distributions. It is straightforward to extend them to multi-dimensional product distributions, corresponding to random vectors with independent components, as in this case the KL is equal to the sum of the KL divergences of the components.
Note that formula \eqref{eq:KL_Laplace} could seem to pose a challenge during gradient-based optimisation due to the presence of the absolute value. However, auto-differentiation packages solve this by calculating left or right derivatives which are defined in every case.

\section{Computing Risk Certificates}
\label{s:kl_invert}

After optimising the distribution over network weights through the previously presented training objectives, we compute a risk certificate on the error of the stochastic predictor, following the procedure of \cite{LangfordCaruana2001}.
This uses the PAC-Bayes-kl bound (\cref{thm:pb-kl}).
First we describe how to invert the binary KL (defined in Eq.~\eqref{eq:binkl}) with respect to its second argument.
For $x\in[0,1]$ and $b\in [0,\infty)$, %let :
we define:
\begin{align*}
f^{\star}(x,b) = \sup \{ y \in [x,1] \,:\, \kl(x \Vert y)\le b \}\,.
\end{align*}
This is easily seen to be well-defined. 
Furthermore, the crucial property that we rely on is that $\kl(x \Vert y)\le b$ holds precisely when $y \leq f^{\star}(x,b)$.

Note that the function $f^{\star}$ provides a way for computing an upper bound on $L(Q)$ based on the PAC-Bayes-kl bound (given in Eq.~\eqref{eq:pb-kl}): For any confidence  $\delta\in(0,1)$, with probability at least $1-\delta$ over size-$n$ random samples $S$ we have:
\begin{align*}
    L(Q)\le f^{\star}\Bigl( \hL_S(Q), \frac{\KL(Q \Vert Q^0)+\log(\frac{ 2\sqrt{n} }{\delta})}{n} \Bigr)\,.
\end{align*}
At this point, as noted by \citet{LangfordCaruana2001}, the difficulty is evaluating $\hL_S(Q)$. This quantity is not computable. Since $f^{\star}$ is a monotonically increasing function of its first argument (when fixing the second argument), it suffices to upper-bound $\hL_S(Q)$.

\subsection{Estimating the Empirical Loss via Monte Carlo Sampling}
 In fact, $f^{\star}$ is also used to estimate the empirical term $\hL_S(Q)$ by random weight sampling: If $W_1,\ldots,W_m \sim Q$ are i.i.d. and $\hat{Q}_m = \sum_{j=1}^{m} \delta_{W_j}$ is the empirical distribution, then for any $\delta'\in(0,1)$, with probability at least $1-\delta'$ we have $\kl(\hL_S(\hat{Q}_m) \Vert \hL_S(Q)) \leq m^{-1}\log(2/\delta')$ (see \citealp[Theorem~2.5]{LangfordCaruana2001}), hence by the inversion formula:
\begin{align*}
    \hL_S(Q)
    \le f^{\star}\Bigl( \hL_S(\hat{Q}_m), \frac{1}{m} \log(\frac{2}{\delta'}) \Bigr)\,.
\end{align*}
This expression can be applied to upper-bound $\hL^{01}_S(Q)$ or $\hL^{\ce}_S(Q)$ by setting the underlying loss function to be the 01 (classification) loss or the cross-entropy loss, respectively. This estimation is valid with high probability (of at least $1-\delta'$) over random weight samples.

The latter expression also can be combined with any of the PAC-Bayes bounds presented in \cref{s:PAC-Bayes} to upper-bound the loss $L(Q_S)$ 
by a computable expression. 
Just to illustrate, combining with the classical PAC-Bayes bound we would get the following risk bound:
\begin{align*}
L(Q_S) 
\leq f^{\star}\Bigl( \hL_S(\hat{Q}_m), \frac{1}{m} \log(\frac{2}{\delta'}) \Bigr) 
+ \sqrt{\frac{\KL(Q_S \Vert Q^0)+\log(\frac{ 2\sqrt{n} }{\delta})}{2n}} \,,
\end{align*}
which holds with probability at least $1-\delta-\delta'$ over random size-$n$ data samples $S$ and size-$m$ weight samples $W_1,\ldots,W_m \sim Q_S$. The parameter $\delta\in(0,1)$ quantifies the confidence over random data samples, and $\delta'\in(0,1)$ the confidence over random weight samples.

As we said before, our evaluation of risk certificates was based on the PAC-Bayes-kl bound. The next subsection fills the details.

\subsection{Final Expression for Evaluating the Risk Certificate}
In our experiments we evaluate the risk certificates (risk upper bounds) for the cross-entropy loss ($\ell^{\ce}$) and the 0-1 loss ($\ell^{01}$), respectively, computed using the PAC-Bayes-kl bound and Monte Carlo weight sampling. For any $\delta,\delta'\in(0,1)$, with probability at least $1-\delta-\delta'$ over random size-$n$ data samples $S$ and size-$m$ weight samples $W_1,\ldots,W_m \sim Q_S$ we have:
\begin{align*}
    L(Q)\le f^{\star}\biggl( f^{\star}\Bigl( \hL_S(\hat{Q}_m), \frac{1}{m} \log(\frac{2}{\delta'}) \Bigr), \frac{\KL(Q \Vert Q^0)+\log(\frac{ 2\sqrt{n} }{\delta})}{n} \biggr)\,.
\end{align*}
In our experiments we used a numerical implementation of the kl inversion $f^{\star}$ and the upper bound just shown to evaluate risk certificates for the stochastic predictors corresponding to the 
distributions over weights obtained by our training methods.

\section{Experimental Results}
\label{s:exp}

We performed a series of experiments on MNIST and CIFAR-10 to thoroughly investigate %
the training objectives presented before
with regards to their ability to give self-certified predictors. 
Specifically, we
 empirically evaluate the two proposed training objectives $f_{\mathrm{quad}}$ and $f_{\mathrm{lambda}}$ of Eq.~\eqref{eq:obj-quad} and Eq.~\eqref{eq:obj-lambda}, and compare these to
 $f_{\mathrm{classic}}$ of Eq.~\eqref{eq:obj-classic} and $f_{\mathrm{bbb}}$ of Eq.~\eqref{eq:bbbobjective}. When possible, we also compare to empirical risk minimisation ($f_{\mathrm{erm}}$) with dropout. 
 In all experiments, training objectives are compared under the same conditions, i.e. weight initialisation, prior, optimiser (vanilla SGD with momentum) and network architecture.
 The code for our experiments is publicly available\footnote{Code available at \url{https://github.com/mperezortiz/PBB}} in PyTorch.

\subsection{Choice of Distribution over Weights}

We studied Gaussian and Laplace distributions over the model weights. The PAC-Bayes posterior distribution $Q$ is learned by optimising a PBB training objective, and is of the same kind as the PAC-Bayes prior (Gaussian or Laplace) in each case.

We also tested in our experiments both data-free random priors (with randomness in the  initialisation of the weights) and data-dependent priors. 
In both cases, the center parameters $\mu_0$ of the prior were initialised randomly from a truncated centered Gaussian distribution with standard deviation set to $1/\sqrt{n_\mathrm{in}}$, where $n_\mathrm{in}$ is the dimension of the inputs to a particular layer, truncating at $\pm 2$ standard deviations. 
The main difference between our data-free and data-dependent priors is that, after initialisation, the center parameters of data-dependent priors are optimised through ERM on a subset of the training data (50\% if not indicated otherwise), while we simply use the initial random weights in the case of data-free priors. 
% After choosing the center parameter of the prior, 
The prior scale parameters $\rho_0$ are set to the constant scale hyper-parameter. 
The posterior $Q$ is always initialised at the prior (both center and scale parameters). 
This means that the posterior center $\mu$ is initialised at the empirical risk minimiser in the case of data-dependent priors, and to the initial random weights in the case of data-free priors. 
We find in our experiments that the prior can be over-fitted easily. To avoid this, we use dropout during the learning process (exclusive to learning the prior, not the posterior).

\subsection{Experimental Setup}

All risk certificates were computed using the the PAC-Bayes-kl inequality, as explained in \cref{s:kl_invert}, with $\delta=0.025$ and $\delta'=0.01$  and $m=150.000$ Monte Carlo model samples, as done by \citet{dziugaite2017computing}. The same confidence $\delta$ was used in all the PBB training objectives ($f_{\mathrm{quad}}$, $f_{\mathrm{lambda}}$, $f_{\mathrm{classic}}$). Input data was standardised.

\subsubsection{Hyperparameter selection} 
\label{sss:hyp_select}
For all experiments we performed a grid search over all hyper-parameters and selected the run with the best risk certificate on 0-1 error\footnote{Note that if we use a total of $C$ hyperparameter combinations, the union bound correction would add no more than $\log(C)/30000$ to the PAC-Bayes-kl upper bound. Even with say $C=$ 42M (forty two million), the value of our risk certificates, computed via kl inversion, will not be impacted significantly. The reader can be assured that we used much less than 42M hyperparameter combinations.} (evaluated as explained in \cref{s:kl_invert}).  
We elaborate more on the use of PAC-Bayes bounds for model selection in the next subsection. We did a grid sweep over the prior distribution scale hyper-parameter (i.e. standard deviation $\sigma_0$) with values in $[0.1, 0.05, 0.04, 0.03, 0.02, 0.01, 0.005]$. 
 We observed that higher variance values lead to instability during training and lower variance does not explore the weight space. 
For the SGD with momentum optimiser we performed a grid sweep over learning rate in $[1\mathrm{e}-3, 5\mathrm{e}-3, 1\mathrm{e}-2]$ and momentum in $[0.95, 0.99]$. We found that learning rates higher than $1\mathrm{e}-2$ caused divergence in training and learning rates lower than $5\mathrm{e}-3$ converged slowly. We also found that the best optimiser hyper-parameters for building the data-dependent prior differ from those selected for optimising the posterior. Because of this, we also performed a grid sweep over the learning rate and momentum used for learning the data-dependent prior (testing the same values as before). The dropout rate used for learning the prior was selected from $[0.0, 0.05, 0.1, 0.2, 0.3]$.
All training objectives derived from PAC-Bayes bounds 
% ($f_{\mathrm{quad}}$, $f_{\mathrm{lambda}}$, $f_{\mathrm{classic}}$) 
used the `bounded cross-entropy' function as surrogate loss during training, for which we enforced boundedness by restricting the minimum probability (see Section~\ref{s:PBB}). We observed that the value $p_\mathrm{min} =1\mathrm{e}-5 $ performed well. Values higher than $1\mathrm{e}-2$ distorts the input to loss function and leads to higher training loss.
The lambda value in $f_{\mathrm{lambda}}$ was initialised to 1.0 (as done by \citealp{thiemann-etal2017}) and optimised using alternate minimisation using SGD with momentum, using the same choice of learning rate and momentum as for the posterior optimisation.
Notice that $f_{\mathrm{bbb}}$ requires an additional sweep over a KL trade-off coefficient, 
which was done with values in $[1\mathrm{e}-5, 1\mathrm{e}-4, \ldots, 1\mathrm{e}-1]$, see \citet{blundell2015weight}. 

For ERM, we used the same range for optimising the learning rate, momentum and dropout rate. However, given that in this case we do not have a risk certificate we need to set aside some data for validation and hyper-parameter tuning. We set 4\% of the data as validation in MNIST (2400 examples) and 5\% in the case of CIFAR-10 (2500 examples). 
At this time we have not done model selection with a risk bound for
this ERM point estimator model, since to the best of our knowledge those bounds are notoriously vacuous and we are not aware of any empirical evidence that they could be used for model selection.

\subsubsection{Predictors and metrics reported}
For all methods, we compare three different prediction strategies using the final model weights: i) stochastic predictor, randomly sampling fresh model weights for each test example; ii) deterministic predictor, using exclusively the posterior mean; iii) ensemble predictor, as done by \citet{blundell2015weight}, in which majority voting is used with the predictions of a number of model weight samples, in our case 100. We report the test cross entropy loss (\ce) and 0-1 error of these predictors. We also report a series of metrics at the end of training (train empirical risk using cross-entropy $\hL^{\ce}_S(Q)$ and 0-1 error $\hL^{01}_S(Q)$ and KL divergence between posterior and prior) and the risk certificate (obtained via PAC-Bayes-kl inversion) for the stochastic predictor ($\ell^{\ce}$ for cross-entropy loss and  $\ell^{01}$ for 0-1 loss).

\subsubsection{Architectures}
For MNIST, we tested both a fully connected neural network (FCN) with 3 layers (excluding the `input layer') and 600 units per hidden layer, as well as a convolutional neural network (CNN) with 4 layers (two convolutional followed by two fully connected). For the latter, we learn a distribution over the convolutional kernels and the weight matrix. 
 We trained our models using the standard MNIST data set split of 60000 training and 10000 test examples. 
For CIFAR-10, we tested three convolutional architectures: one with a total of 9 layers with learnable parameters and the other two with 13 and 15 layers; and we used the standard data set split of 50000 training and 10000 test examples.
ReLU activations were used in each hidden layer for both data sets.
Both for learning the posterior and the prior, we ran the training for 100 epochs (however we observed that methods converged around 70). We used a training batch size of $250$ for all the experiments.

\begin{figure}[ht]
\centering
\includegraphics[width=\textwidth]{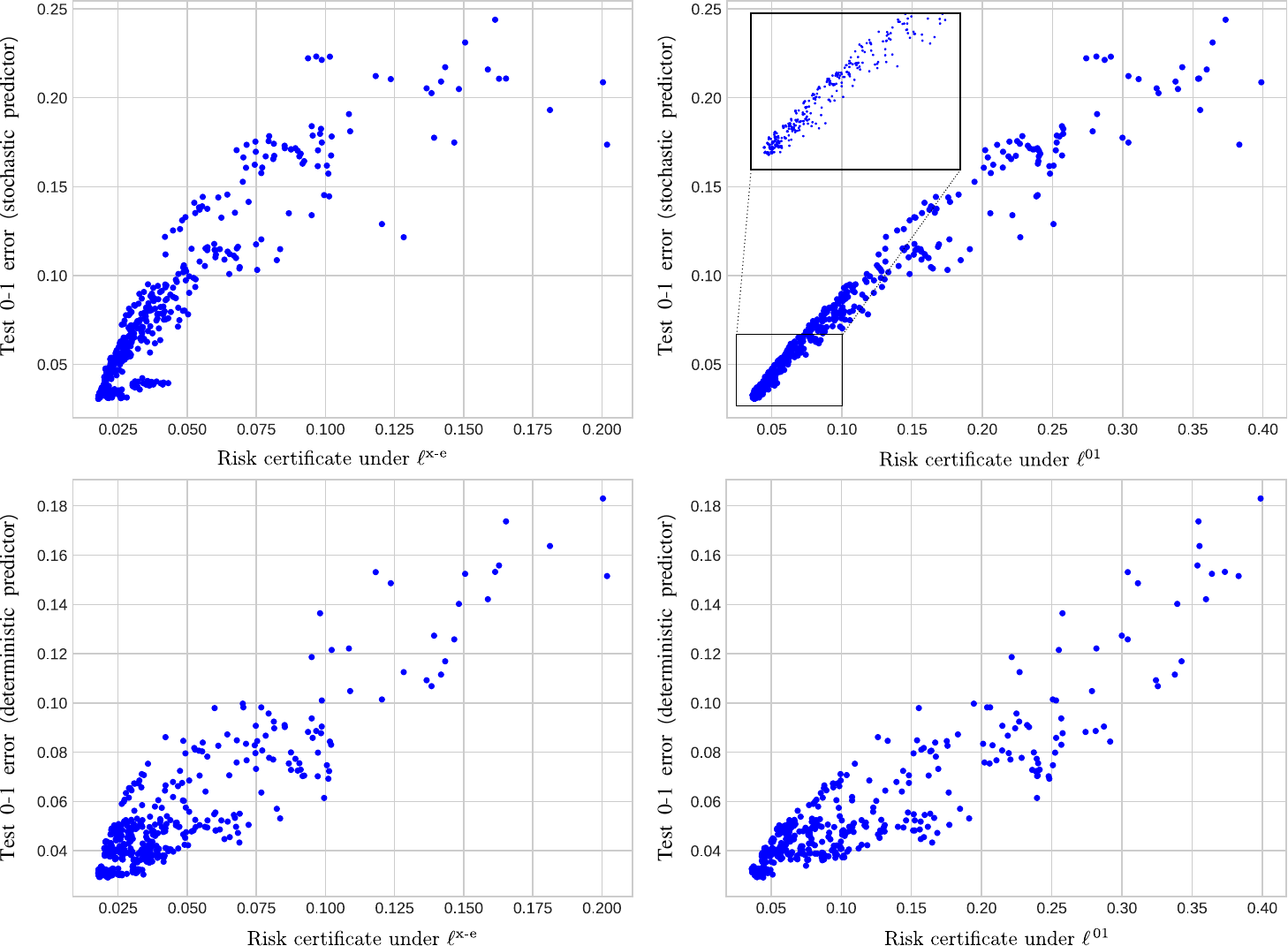}
\caption{Model selection results from more than 600 runs with different hyper-parameters. 
We use a reduced subset of MNIST for these experiments (10\% of training data). The architecture used is a CNN, with Gaussian distributions over weights and data-dependent PAC-Bayes priors. 
The plots show the values of risk certificates (under $\ell^{\ce}$ and $\ell^{01}$) on the horizontal axes, and the test set error rates on the vertical axes, for both the stochastic and deterministic predictors.}
\label{fig:gridsearch}
\end{figure}

\subsection{Hyper-parameter and Architecture Search through PAC-Bayes Bounds}

We show now that PAC-Bayes bounds can be used not only as training objectives to guide 
the optimisation algorithm but also for model selection. Specifically, \figurename{ \ref{fig:gridsearch}} compares 
the PAC-Bayes-kl bound for cross-entropy and 0-1 %error
losses (x-axis) to the test 0-1 error for the stochastic predictor (y-axis, top row) and deterministic predictor (y-axis, bottom row) for more than 600 runs from the hyper-parameter grid search performed for $f_{\mathrm{quad}}$ with a CNN architecture and a data-dependent Gaussian prior on MNIST. 
We do a grid search over 6 hyper-parameters: prior scale, dropout rate, and the learning rate and momentum both for learning the prior and the posterior. To depict a larger range of performance values (thus avoiding only showing the risk and performance for relatively accurate classifiers) we use here a reduced training set for these experiments (i.e. 10\% of training data from MNIST). The test set is  maintained. The results show a clear positive correlation between the risk certificate and test set 0-1 error of the stochastic predictor, especially for the risk certificate of the 0-1 error, as expected. The results are obviously not as positive for the test 0-1 error of the deterministic predictor (since the bound is on the stochastic predictor), but there still exist a linear trend. While the plots also show heteroskedasticity (there is a noticeable increase of variability towards the right side of the x-axis) the crucial observation is that for small error values the corresponding values of the risk certificate are reasonable stable. It is worth keeping in mind, however, that bounds generally get weaker with higher error values.

\begin{figure}[htb]
\centering
\includegraphics[width=8.5cm]{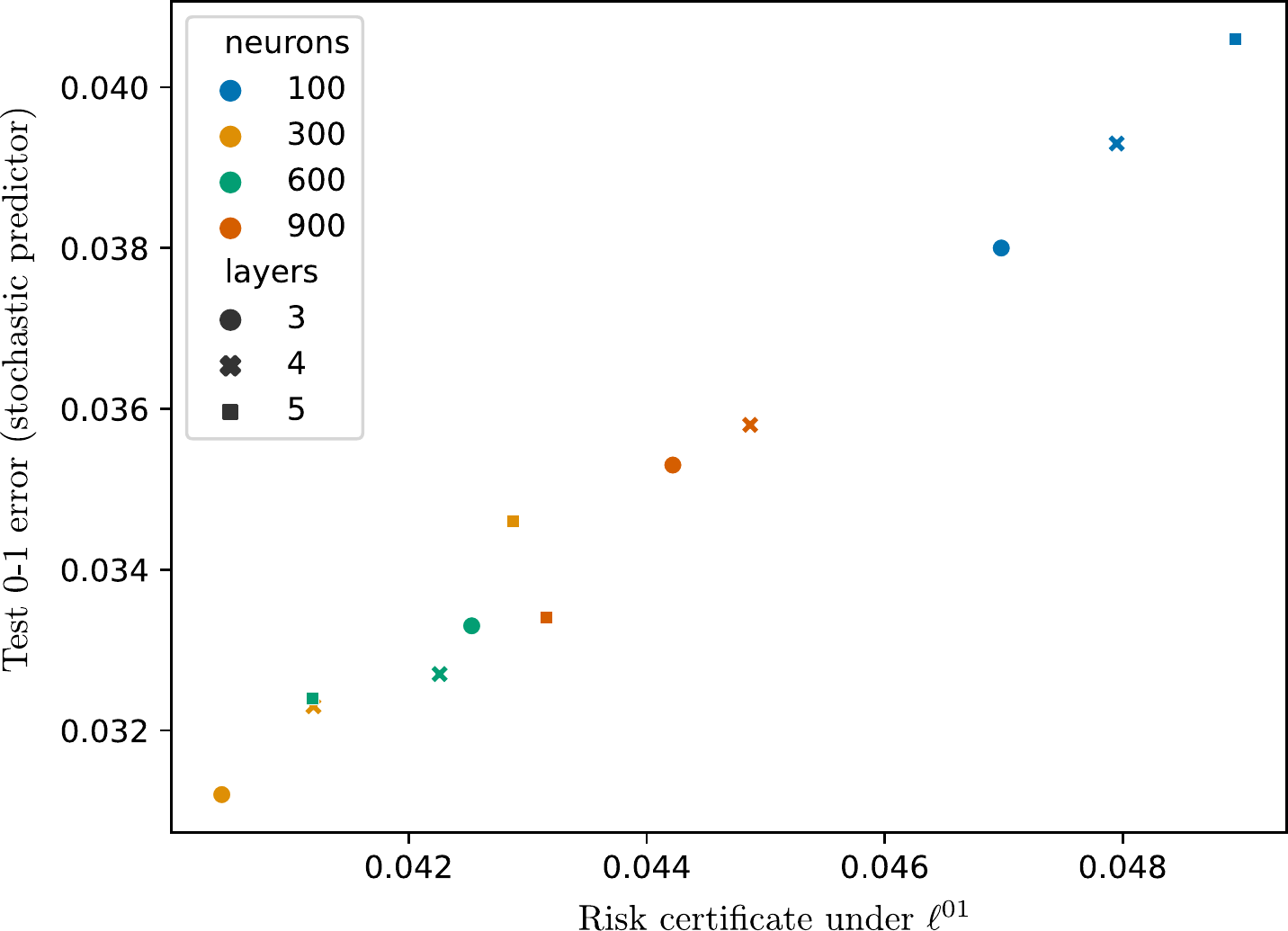}
\caption{Risk certificate under $\ell^{01}$ vs test 0-1 error on MNIST for a set of fully connected architectures (varying the number of layers and number of neurons per layer).}
\label{fig:archsearch}
\end{figure}

\figurename{ \ref{fig:archsearch}} shows a different experiment regarding model selection using MNIST and a fully connected architecture. In this case, we fix the hyperparameters and run several versions of the network with different number of layers and neurons per layer. All the networks are trained in the exact same way using $f_{\mathrm{quad}}$. The linear trend between the risk certificate under $\ell^{01}$ and the test 0-1 error further validates the usefulness of the risk certificate under $\ell^{01}$ for model selection.

Motivated by the results %i
shown in \figurename{ \ref{fig:gridsearch}} and \figurename{ \ref{fig:archsearch}},
where it is shown that the bound could potentially be used for model selection, we use the risk certificate % 
with $\ell^{01}$ (evaluated as explained in \cref{s:kl_invert}) for hyper-parameter tuning in all our subsequent experiments. Note that the advantage in this case is that our approach obviates the need of
a held-out set of examples
for hyper-parameter tuning.

\begin{table}[htb]
\centering
% \begin{center}
\setlength{\tabcolsep}{2.1pt}
\scriptsize
\renewcommand{\arraystretch}{1.0}
 \begin{tabular}{|c|c|c| c c | c c c | c c | c c | c c | c |} 
 \hline
 \multicolumn{3}{|c}{Setup }& \multicolumn{5}{|c}{Risk cert. \ \& \quad Train metrics \quad\quad} & \multicolumn{2}{|c}{Stch. pred.} & \multicolumn{2}{|c}{Det. pred.} & \multicolumn{2}{|c}{Ens. pred.} & \multicolumn{1}{|c|}{Prior}\\ \hline
\rule{0pt}{8pt}
Arch. & Prior & Obj. & $\ell^{\ce}$ & $\ell^{01}$ & KL/n & $\hL^{\ce}_S(Q)$ & $\hL^{01}_S(Q)$ & \ce & 01 err. & \ce &	 01 err.&	 \ce &	 01 err.&	 01 err.\\ \hline 
 \multirow{9}{19pt}{\centering FCN} &\multirow{4}{45pt}{\centering Rand.Init. (Gaussian)}&  $f_{\mathrm{quad}}$ & .2033 & \textbf{.3155} & .1383 & .0277  & .0951 & .0268 & .0921 & .0137 & .0558 & .0007 & .0572 & .8792\\
& &  $f_{\mathrm{lambda}}$ & .2326 & \emph{.3275} & .1856 & .0218 & .0742 & .0211 & \emph{.0732} & .0077 & .0429 & .0004 & .0448 & .8792\\
& & $f_{\mathrm{classic}}$ & .1749 & .3304 & .0810 & .0433 & .1531 & .0407 & .1411 & .0204 & .0851 & .0009 & .0868 & .8792\\
& & $f_{\mathrm{bbb}}$ & .5163 & .5516 & .6857 & .0066 & .0235 & .0088 & \textbf{.0293} & .0038 & .0172 & .0003 & .0178 & .8792\\ 
\cline{2-15}
 & \multirow{4}{45pt}{\centering Learnt  (Gaussian)}&  $f_{\mathrm{quad}}$ & .0146 & \textbf{.0279} & .0010 & .0092 & .0204 & .0084 & .0202 & .0032 & .0186 & .0002 & .0189 & .0202 \\
& & $f_{\mathrm{lambda}}$ & .0201 & .0354 & .0054 & .0073 & .0178 & .0082 & \emph{.0196} & .0071 & .0185 & .0001 & .0185 & .0202 \\
& & $f_{\mathrm{classic}}$ & .0141 & \emph{.0284} & .0001 & .0115 & .0247 & .0101 & .0230 & .0089 & .0189 & .0002 & .0191 & .0202 \\
& & $f_{\mathrm{bbb}}$ & .0788 & .0968 & .0704 & .0025 & .0090 & .0063 & \textbf{.0179} & .0066 & .0153 & .0001 & .0153 & .0202\\ %\hline 
\cline{2-15}
& - & $f_{\mathrm{erm}}$ & - & - & - & .0004 & .0007 & - & - & .0101 & .0152 & - & - & - \\ 
 \hline %\hline
 \multirow{9}{19pt}{\centering  CNN} & \multirow{4}{45pt}{\centering  Rand.Init.  (Gaussian)}&  $f_{\mathrm{quad}}$ & .1453 & \textbf{.2165} & .1039 & .0157 & .0535 & .0143 & .0513 & .0062 & .0257 & .0003 & .0261 & .9478\\
& & $f_{\mathrm{lambda}}$ & .1583 & \emph{.2202} & .1256 & .0126 & .0430 & .0109 & \emph{.0397} & .0056 & .0207 & .0003 & .0211 & .9478\\
&  & $f_{\mathrm{classic}}$ & .1260 & .2277 & .0622 & .0273 & .0932 & .0253 & .0869 & .0111 & .0425 & .0006 & .0421 & .9478\\
& & $f_{\mathrm{bbb}}$ & .3400 & .3645 & .3948 & .0034 & .0120 & .0039 & \textbf{.0154} & .0016 & .0088 & .0001 & .0092 & .9478\\ \cline{2-15}
% \hline 
& \multirow{4}{45pt}{\centering Learnt  (Gaussian)}&  $f_{\mathrm{quad}}$ & .0078 & \textbf{.0155} & .0001 & .0058 & .0127 & .0045 & \textbf{.0104} & .0003 & .0105 & .0001 & .0104 & .0104 \\
& & $f_{\mathrm{lambda}}$ & .0095 & .0186 & .0010 & .0051 & .0123 & .0044 & \emph{.0106} & .0047 & .0098 & .0000 & .0100 & .0104 \\
& & $f_{\mathrm{classic}}$ & .0083 & \emph{.0166} & .0000 & .0064 & .0139 & .0049 & .0123 & .0048 & .0103 & .0001 & .0103 & .0104 \\
& & $f_{\mathrm{bbb}}$ & .0447 & .0538 & .0398 & .0012 & .0042 & .0040 & \textbf{.0104} & .0043 & .0082 & .0002 & .0082 & .0104 \\ \cline{2-15}
 & - & $f_{\mathrm{erm}}$ & - & - & - & .0003 & .0004 & - & - & .0081 & .0092 & - & - & - \\ %\hline
 \hline
\end{tabular}
% \end{center}
\caption{Training and test set metrics on MNIST using Gaussian distributions over weights. The table includes two architectures (FCN and CNN), two kinds of PAC-Bayes priors (a data-free prior centered at the randomly initialised weights, and a data-dependent prior learnt on a subset of the data set) and four training objectives. 
For the stochastic predictor, the best risk certificate and test set error are highlighted in bold face, and second best are highlighted in italics. } 
\label{tab:mnist_big_table}
\end{table}

\begin{figure}[htb]
\centering
\includegraphics[width=\textwidth]{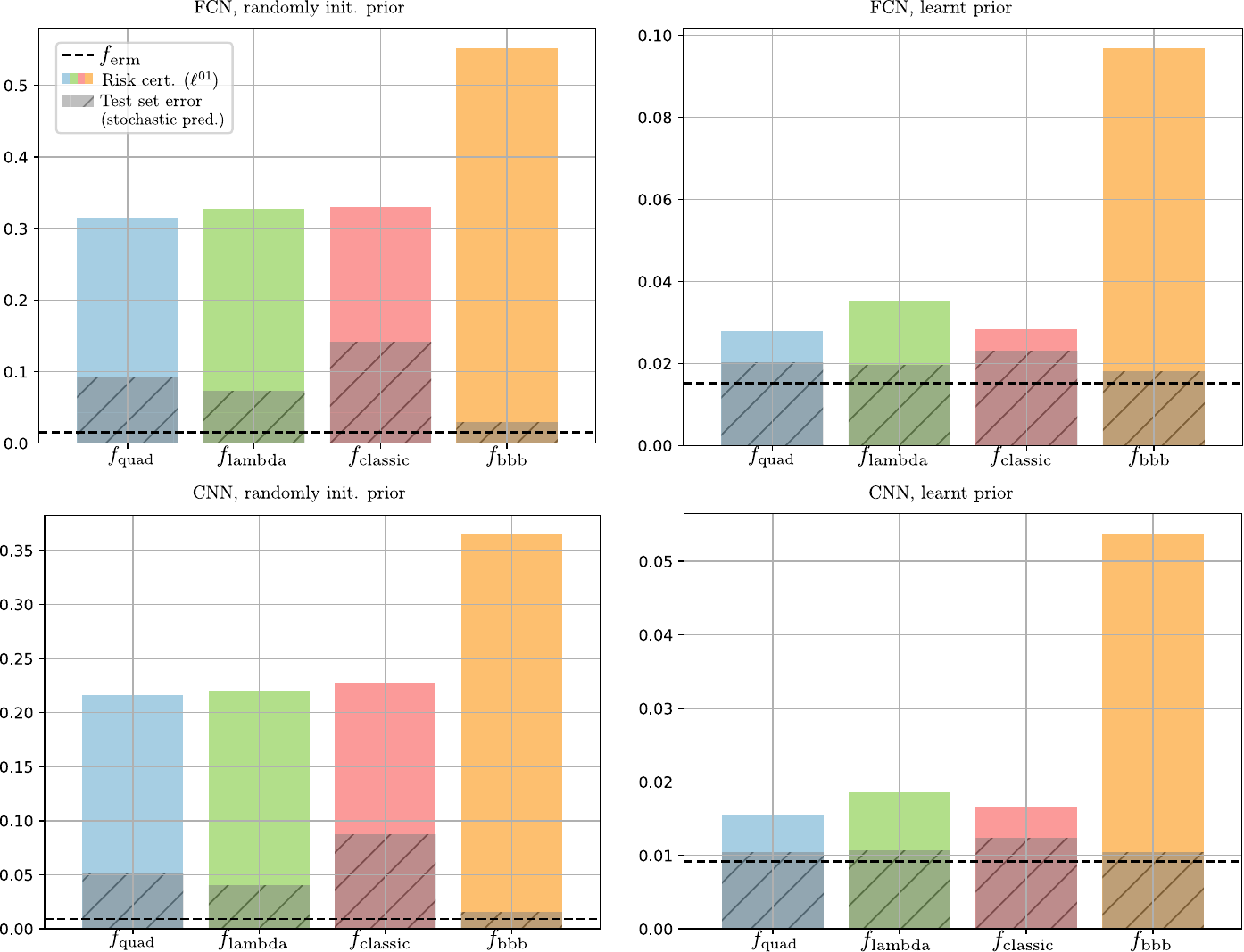}
\caption{ 
% Bar plots of results across different architectures, priors and training objectives for MNIST. 
Tightness of the risk certificates for MNIST across different architectures, priors and training objectives.
The bottom shaded areas correspond to the test set 0-1 error of the stochastic classifier. 
The coloured areas on top indicate the tightness of the risk certificate (smaller is better). 
The horizontal dashed line corresponds to the test set 0-1 error of $f_{\mathrm{erm}}$, i.e. the deterministic classifier learnt by empirical risk minimisation of the surrogate loss on the whole training set (shown for comparison purposes).}
\label{fig:mnistbar}
\end{figure}

\subsection{Comparison of Different Training Objectives and Priors}
\label{exp_MNIST}

We first present a comparison of the four considered training objectives on MNIST using Gaussian distributions over weights. Table \ref{tab:mnist_big_table} shows the results for the two architectures previously described for MNIST (FCN and CNN) and both data-free and data-dependent priors (referred to as Rand.Init. and Learnt, respectively). We also include the results obtained by standard ERM using the cross-entropy loss, for which part of the table can not be completed (e.g. risk certificates). The last column of the table shows the test set 0-1 error of the prior mean deterministic predictor (column named Prior). We also report the test set performance for the stochastic predictor (Stch. pred.), the posterior mean deterministic predictor (Det. pred.) and the ensemble predictor (Ens pred.). 
For all the reported results and tables, we highlight the best risk certificate and stochastic test set error in bold face and the second best is highlighted in italics.

An important note is that we used the risk certificates for model selection for all training objectives, including $f_{\mathrm{bbb}}$ (with the sole exception of $f_{\mathrm{erm}}$, for which we used a validation set due to the reasons discussed in~\cref{sss:hyp_select}). The KL trade-off coefficient included in $f_{\mathrm{bbb}}$ \citep{blundell2015weight} relaxes the importance given to the prior in the optimisation, but obviously not in the computation of the risk certificate, which in practice means that larger KL attenuating coefficients will lead to worse risk certificates. Because of this, in all cases, the model selection strategy chose the lowest value (namely, 0.1) for the KL attenuating coefficient  for $f_{\mathrm{bbb}}$, meaning there are cases in which $f_{\mathrm{bbb}}$ obtained better test set performance than the ones we report in this table, but much looser risk certificates. We present more experiments on this in the next subsection where we experiment with the KL attenuating trick.

The findings from our experiments on MNIST, reported in \cref{tab:mnist_big_table} and \figurename{ \ref{fig:mnistbar}}, are as follows: i) $f_{\mathrm{quad}}$ achieves consistently the best risk certificates for 0-1 error (see $\ell^{01})$ in all experiments, providing as well better test performance than $f_{\mathrm{classic}}$, as observed when comparing the 0-1 loss of the stochastic predictors. ii) Based on the results of the stochastic predictor, $f_{\mathrm{lambda}}$ is the best PAC-Bayes inspired objective in terms of test performance, although the risk certificates are generally less tight.  iii) In most cases, the stochastic predictor does not worsen the performance of the prior mean predictor, improving it very significantly for random data-free priors (i.e. Rand.Init). iv) The mean of the weight distribution is also improved, as shown by comparing the results of the deterministic predictor (Det. pred.),
corresponding to the posterior mean,
with the prior mean predictor. The ensemble predictor also generally improves on the prior. v) The improvements brought by data-dependent priors (labelled as ``Learnt'' in the table) are consistent across the two architectures, showing better test performance and risk certificates (although the use of data-free priors still produced non-vacuous risk certificates). vi) The application of PBB is successful not only for learning fully connected layers but also for learning convolutional ones. The improvements in performance and risk certificates that the use of a CNN brings are also noteworthy. vii) The proposed PAC-Bayes inspired learning strategies show competitive performance (specially when using data-dependent priors) when compared to 
the Bayesian inspired
$f_{\mathrm{bbb}}$ and the widely-used $f_{\mathrm{erm}}$. Besides this comparable test set performance, our training methods also provide risk certificates with tight values.

We now compare our results to those reported 
by \citet{dziugaite2018data} for MNIST.
%% before in the PAC-Bayes literature for MNIST. 
Note that in this case there are differences regarding optimiser, prior chosen and weight initialisation (however, the neural network architecture used is the same, FCN as described in this paper).
\citet{dziugaite2018data} evaluated the bound of their Theorem 4.2 and the bound of \citet{lever-etal2013} for comparison.
We compare the results reported by them with the results of training with our two training objectives $f_{\mathrm{quad}}$ and $f_{\mathrm{lambda}}$, and with $f_{\mathrm{classic}}$ (optimised as per our $f_{\mathrm{quad}}$ and $f_{\mathrm{lambda}}$). These results are presented in \cref{tab:mnist}.
\begin{table}[htb]
\centering
% \begin{center}
\footnotesize
\renewcommand{\arraystretch}{1.1}
\begin{tabular}{| c| c| c| c| c|} 
 \hline
& Training method & Stch. Pred. 01 Err & Risk cert. $\ell^{01}$ & Bound used \\
\hline
\hline
\multirow{2}{*}{D{\&}R 2018} 
    & SGLD & \multirow{2}{*}{0.1200} & 0.2100 & D\&R18 Thm. 4.2 \\[-1mm]
    & ($\tau=3\mathrm{e}+3$) &  & 0.2600 & Lever et al. 2013 \\ 
 \hline 
\multirow{2}{*}{D{\&}R 2018}  
    & SGLD & \multirow{2}{*}{0.0600} & 0.6500 & D\&R18 Thm. 4.2 \\[-1mm]
    & ($\tau=1\mathrm{e}+5$) &  & 1.0000 & Lever et al. 2013 \\ 
 \hline
\multirow{3}{*}{This work}
    & SGD + $f_{\mathrm{quad}}$ & \emph{0.0202} & \textbf{0.0279} & PAC-Bayes-kl \\
    & SGD + $f_{\mathrm{lambda}}$ &  \textbf{0.0196} &  0.0354 & PAC-Bayes-kl \\
    & SGD + $f_{\mathrm{classic}}$ &  0.0230 & \emph{0.0284} & PAC-Bayes-kl \\
 \hline
\end{tabular}
% \end{center}
\caption{Comparison of test set error rate (0-1 loss) for the stochastic predictor and its risk certificate for standard MNIST data set. We compare here our results for the FCN with data-dependent priors to previous published work. All methods use data-dependent priors (albeit different ones) and exactly the same architecture of dimensions
$784 \times 600 \times 600 \times 10$
(with 2 hidden layers of 600 units per layer).}
\label{tab:mnist}
\end{table}

The hyperparameter $\tau$ in both \citet{dziugaite2018data} and \citet{lever-etal2013} controls the temperature of a Gibbs distribution with unnormalised density $e^{-\tau\hL_S(w)}$ with respect to some fixed measure on weight space.  
In the table we display only the two values of their $\tau$ parameter which achieve best test set error and risk certificate.
We note that the best values reported by \citet{dziugaite2018data} correspond to test accuracy of 94\% or 93\% while in those cases their risk certificates (0.650 or 0.350, respectively), although non-vacuous, were far from being tight. On the other hand, the tightest value of their risk bound (0.21) only gives an 88\% accuracy.
In contrast, our PBB methods achieve close to 98\% test accuracy (or 0.0202 test error).
At the same time, as noted above, our risk certificate (0.0279) is much tighter than theirs (0.210), meaning that our training scheme (not only training objectives but also prior) are a significant improvement with respect to theirs (an order of magnitude tighter). Even more accurate predictors and tighter bounds are achieved by the CNN architecture, as shown in \cref{tab:mnist_big_table}.

\subsection{KL Attenuating Trick}

As many works have pointed out before (and we have observed in our experiments), the problem with all the four presented training objectives is that the KL term tends to dominate and most of the work in training is targeted at reducing it, which effectively means often the posterior cannot move far from the prior. 
To address this issue, distribution-dependent \citep{lever-etal2013} or data-dependent \citep{dziugaite2018data} priors have been used in the literature. Another approach to address this is to add a coefficient that controls the influence of the KL in the training objective \citep{blundell2015weight}.
This means that in the case of $f_{\mathrm{bbb}}$ we could see marginal decrease in the KL divergence during the course of training (specially given small KL attenuating coefficients) and the solution it returns is expected to be similar to that returned simply using ERM with cross-entropy. 
However, this also has its effects on the risk certificate. To show these effects, we run all four training objectives with a KL penalty of 0.0001 during training and report the results in \cref{tab:mnist_penalty}. For simplicity, only a CNN architecture is considered in this experiment. What we can see comparing these results to the ones reported in \cref{tab:mnist_big_table} is that while the 0-1 error for the stochastic classifier decreases, the KL term increases and so does the final risk certificate. Practitioners may want to consider this trade-off between test set performance and tightness of the risk certificates.

\begin{table}[t]
\centering
% \begin{center}
\setlength{\tabcolsep}{2.5pt}
\scriptsize
\renewcommand{\arraystretch}{1.1}
 \begin{tabular}{|c|c|  c c |  c c c | c c | c c | c c | c |} 
 \hline
 \multicolumn{2}{|c}{Setup }& \multicolumn{5}{|c}{Risk cert. \ \& \quad Train metrics \quad\quad} & \multicolumn{2}{|c}{Stch. pred.} & \multicolumn{2}{|c}{Det. pred.} & \multicolumn{2}{|c}{Ens. pred.} & \multicolumn{1}{|c|}{Prior}\\ \hline
\rule{0pt}{8pt}
Arch. \& Prior & Obj. & $\ell^{\ce}$ & $\ell^{01}$ & KL/n & $\hL^{\ce}_S(Q)$ & $\hL^{01}_S(Q)$ & \ce & 01 err. & \ce &	 01 err.&	 \ce &	 01 err.&	 01 err.\\ \hline 
 \multirow{4}{52pt}{\centering CNN \\ Rand.Init \\ (KL attenuating)}&  $f_{\mathrm{quad}}$ & .2292 & \textbf{.2824} & .2174 & .0097 & .0330 & .0084 & .0305 & .0042 & .0193 & .0002 & .0201 & .9478 \\
&  $f_{\mathrm{lambda}}$ & .2840 & .3241 & .3004 & .0066 & .0225 & .0058 & \emph{.0222} & .0039 & .0144 & .0002 & .0148 & .9478 \\
&  $f_{\mathrm{classic}}$ & .2297 & \emph{.2846} & .2167 & .0101 & .0344 & .0096 & .0343 & .0047 & .0208 & .0002 & .0216 & .9478 \\
&  $f_{\mathrm{bbb}}$ & .4815 &	.4974 &	.6402 &	.0024 &	.0082 &	.0035 &	\textbf{.0107} &	.0024 &	.0082 &	.0000 &	.0079 &	.9478 \\ \hline 
 \multirow{4}{45pt}{\centering CNN \\ Learnt \\ (KL attenuating)}&  $f_{\mathrm{quad}}$ & .0191 & \textbf{.0296} & .0104 & .0030 & .0087 & .0033 & .0101 & .0000 & .0095 & .0000 & .0096 & .0104 \\
&  $f_{\mathrm{lambda}}$ & .0245 & \emph{.0354} & .0162 & .0025 & .0076 & .0031 & \textbf{.0092} & .0040 & .0092 & .0000 & .0095 & .0104 \\
&  $f_{\mathrm{classic}}$ & .0187 & \textbf{.0296} & .0100 & .0031 & .0089 & .0037 & .0106 & .0043 & .0095 & .0001 & .0095 & .0104 \\
&  $f_{\mathrm{bbb}}$ & .0470 & .0557 & .0421 & .0012 & .0041 & .0034 & \emph{.0096} & .0025 & .0085 & .0001 & .0083 & .0104 \\ \hline
\end{tabular}
% \end{center}
\caption{Training and test set results on MNIST using Gaussian distributions over weights and a penalty of $\eta=0.001$ on the KL term for all the training objectives shown. Only a CNN architecture is considered.} 
\label{tab:mnist_penalty}
\end{table}

\subsection{Laplace Weight Distributions}

We experimented with both 
Laplace and Gaussian %priors. 
distributions over weights.
The results are presented in \cref{tab:mnist_laplace}. Comparing these to the results with Gaussian weight distributions from \cref{tab:mnist_big_table}, we did not observe significant and consistent differences
in terms of risk certificates and test set error between the two kinds of prior/posterior distributions. 
The distribution to use could be problem-dependent, but we found that both Gaussian and Laplace distributions achieve good risk certificates and test set performance.

\begin{table}[t]
\centering
% \begin{center}
\setlength{\tabcolsep}{2.3pt}
\scriptsize
\renewcommand{\arraystretch}{1.0}
 \begin{tabular}{|c|c|c| c c | c c c | c c | c c | c c | c |} 
 \hline
 \multicolumn{3}{|c}{Setup }& \multicolumn{5}{|c|}{Risk cert. \ \& \quad Train metrics \quad\quad} & \multicolumn{2}{|c}{Stch. pred.} & \multicolumn{2}{|c}{Det. pred.} & \multicolumn{2}{|c}{Ens. pred.} & \multicolumn{1}{|c|}{Prior}\\ \hline
\rule{0pt}{8pt}
Arch. & prior & Obj. & $\ell^{\ce}$ & $\ell^{01}$ & KL/n & $\hL^{\ce}_S(Q)$ & $\hL^{01}_S(Q)$ & \ce & 01 err. & \ce &	 01 err.&	 \ce &	 01 err.&	 01 err.\\ \hline 
\multirow{8}{19pt}{\centering CNN}&
 \multirow{4}{40pt}{\centering Rand.Init. (Laplace)}&  $f_{\mathrm{quad}}$ & .1548 & \textbf{.2425} & .1024 & .0207 & .0709 & .0190 & .0677 & .0113 & .0429 & .0004 & .0436 & .9478\\
& & $f_{\mathrm{lambda}}$ & .1844 & .2540 & .1489 & .0147 & .0496 & .0131 & \emph{.0461} & .0096 & .0310 & .0003 & .0312 & .9478\\
& & $f_{\mathrm{classic}}$ & .1334 & \emph{.2489} & .0610 & .0322 & .1101 & .0296 & .1014 & .0208 & .0719 & .0007 & .0695 & .9478\\
& & $f_{\mathrm{bbb}}$ & .4280 & .4487 & .5385 & .0031 & .0107 & .0038 & \textbf{.0139} & .0006 & .0096 & .0001 & .0090 & .9478\\\cline{2-15} 
& \multirow{4}{40pt}{\centering Learnt (Laplace)}&  $f_{\mathrm{quad}}$ & .0085 & \emph{.0167} & .0004 & .0056 & .0126 & .0043 & \emph{.0098} & .0011 & .0103 & .0001 & .0103 & .0104 \\
& & $f_{\mathrm{lambda}}$ & .0119 & .0216 & .0025 & .0049 & .0118 & .0041 & .0106 & .0052 & .0103 & .0003 & .0100 & .0104 \\
& & $f_{\mathrm{classic}}$ & .0076 & \textbf{.0155} & .0000 & .0060 & .0131 & .0046 & .0107 & .0015 & .0105 & .0001 & .0106 & .0104 \\
& & $f_{\mathrm{bbb}}$ & .0737 & .0866 & .0673 & .0019 & .0062 & .0031 & \textbf{.0092} & .0013 & .0093 & .0001 & .0091 & .0104 \\ \hline
\end{tabular}
% \end{center}
\caption{Training and test set results on MNIST using Laplace  distributions over weights. For simplicity, only a CNN architecture is considered here.} 
\label{tab:mnist_laplace}
\end{table}

\figurename{ \ref{fig:mnistplotall}} shows a summary of all the results obtained for MNIST (i.e. results reported in \cref{tab:mnist_big_table} and \cref{tab:mnist_laplace}). This shows clearly the differences between the three training objectives: 
\begin{figure}[htb]
\centering
\includegraphics[width=9cm]{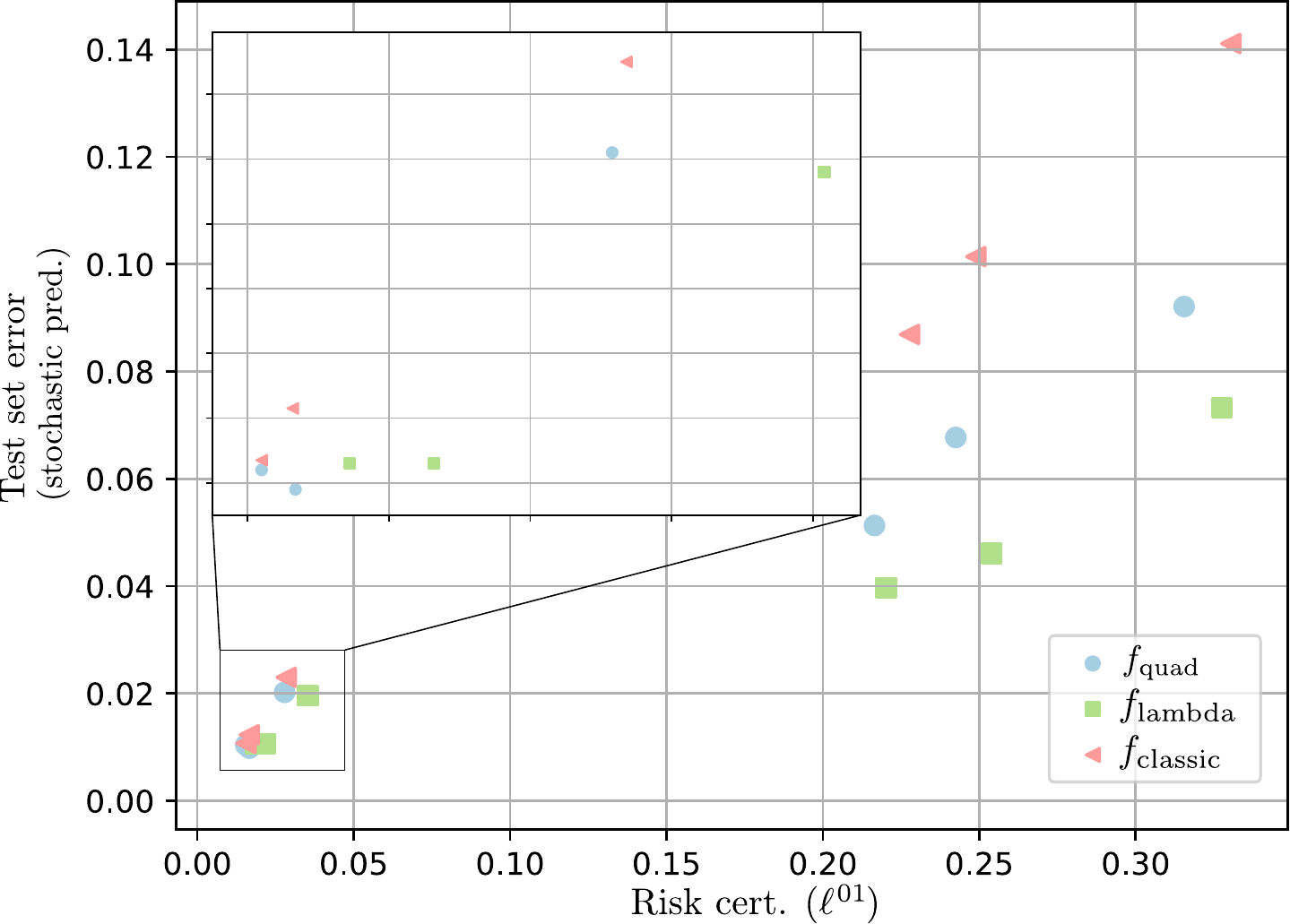}
\caption{Scatter plot of the results obtained for MNIST using different training objectives. 
The x-axis shows values of the risk certificate 
(under $\ell^{01}$ loss),
and the y-axis shows the test set error rates, 
achieved by the stochastic classifier.}
\label{fig:mnistplotall}
\end{figure}
$f_\mathrm{lambda}$ tends to lead generally to the lowest test set error, but worse risk certificates than $f_\mathrm{quad}$, and $f_\mathrm{classic}$ leads to the worse test set performance and looser bounds. Thus, $f_\mathrm{quad}$ gives a reasonable trade-off between test set performance and tight risk certificates. The general trend of the relationship shows a slight curvature, as also seen in \figurename{ \ref{fig:gridsearch}}.

\subsection{CIFAR-10 with Larger Architectures}

\begin{table}[htb]
\centering
% \begin{center}
\setlength{\tabcolsep}{2.3pt}
\scriptsize
\renewcommand{\arraystretch}{1.0}
 \begin{tabular}{|c|c|c| c c | c c c | c c | c c | c c | c |} 
 \hline
 \multicolumn{3}{|c}{Setup }& \multicolumn{5}{|c|}{Risk cert. \ \& \quad Train metrics \quad\quad} & \multicolumn{2}{|c}{Stch. pred.} & \multicolumn{2}{|c}{Det. pred.} & \multicolumn{2}{|c}{Ens. pred.} & \multicolumn{1}{|c|}{Prior}\\ \hline
\rule{0pt}{8pt}
Arch. & Prior & Obj. & $\ell^{\ce}$ & $\ell^{01}$ & KL/n & $\hL^{\ce}_S(Q)$ & $\hL^{01}_S(Q)$ & \ce & 01 err. & \ce &	 01 err.&	 \ce &	 01 err.&	 01 err.\\ \hline  
 \multirow{9}{30pt}{\centering CNN \\ (9 layers)}& \multirow{4}{30pt}{\centering Learnt (50\% data)}&  $f_{\mathrm{quad}}$ &  .1296 & \emph{.3034} & .0089 & .0868 & .2428 & .0903 & .2452 & .0726 & .2439 & .0024 & .2413 & .2518 \\
& & $f_{\mathrm{lambda}}$ & .1742 &	.3730 &	.0611 &	.0571 &	.2108 &	.0689 &	\emph{.2307} &	.0609 &	.2225 &	.0018	& .2133 &	.2518 \\
& & $f_{\mathrm{classic}}$ & .1173 &	\textbf{.2901} &	.0035 &	.0903 &	.2511 &	.0931 &	.2537 &	.0952 &	.2437 &	.0025 &	.2332 &	.2518\\
& & $f_{\mathrm{bbb}}$ & .8096 & .8633 & 1.5107 & .0239 & .0926 & .0715 & \textbf{.2198} & .0735 & .2160 & .0017 & .2130 & .2518 \\ \cline{2-15}
& \multirow{4}{30pt}{\centering Learnt (70\% data)}&  $f_{\mathrm{quad}}$ &  .1017 & \emph{.2502} & .0026 & .0796 & .2179 & .0816 & \emph{.2137} & .0928 & {.2137} & .0023 & .2100 & .2169 \\
& & $f_{\mathrm{lambda}}$ &.1414 & .3128 & .0307 & .0630 & .2022 & .0708 & \textbf{.2081} & .0767 & .2061 & .0021 & .2049 & .2169 \\
& & $f_{\mathrm{classic}}$  & .0957 &  \textbf{.2377} &  .0004 &  .0851 &  .2223 &  .0862 &  .2161 &  .0827 &  .2167 &  .0021 &  .2135 &  .2169  \\
& & $f_{\mathrm{bbb}}$ & .6142 &  .6965 &  .8397 &  .0212 &  .0822 &  .0708 &  .1979 &  .0562 &  .1992 &  .0019 &  .1944 &  .2169 \\\cline{2-15}
& - & $f_{\mathrm{erm}}$ & - & - & - & .0355 & .0552 & - & - & .1400 & .1946 & - & - & - \\ 
\hline
\multirow{9}{30pt}{\centering CNN \\ (13 layers)}& \multirow{4}{30pt}{\centering Learnt (50\% data)}&  $f_{\mathrm{quad}}$ & .0821 & \emph{.2256} & .0042 & .0577 & .1874 & .0585 & .1809 & .0519 & .1788 & .0011 & .1783 & .1914 \\
& & $f_{\mathrm{lambda}}$ & .1163 & .2737 & .0272 & .0491 & .1741 & .0516 & \emph{.1740} & .0466 & .1726 & .0015 & .1690 & .1914 \\
& & $f_{\mathrm{classic}}$ & .0757 & \textbf{.2127} & .0009 & .0635 & .1936 & .0622 & .1880 & .0592 & .1810 & .0017 & .1816 & .1914 \\
& & $f_{\mathrm{bbb}}$ & .6787 & .7566 & .9999 & .0250 & .0924 & .0505 & \textbf{.1676} & .0422 & .1646 & .0011 & .1614 & .1914 \\  \cline{2-15}
& \multirow{4}{30pt}{\centering Learnt (70\% data)}&  $f_{\mathrm{quad}}$ & .0659 & \emph{.1832} & .0015 & .0519 & .1608 & .0517 & .1568 & .0421 & .1553 & .0010 & .1546 & .1587\\
& & $f_{\mathrm{lambda}}$ & .0896 & .2177 & .0145  & .0449  & .1499  & .0479  &  \emph{.1541}  &  .0604  &  .1522  &  .0011  &  .1507  & .1587 \\
& & $f_{\mathrm{classic}}$ & .0619 & \textbf{.1758} & .0002 & .0548 & .1644 & .0541 & .1588 & .0605 & .1578 & .0013 & .1557 & .1587 \\
& & $f_{\mathrm{bbb}}$ & .4961 &  .5858 &  .5826 &  .0213 &  .0772 &  .0487 &  \textbf{.1508} &  .0532 &  .1495 &  .0016 &  .1461 &  .1587 \\  \cline{2-15}
& - & $f_{\mathrm{erm}}$ & - & - & - & .0576 & .0810 & - & - & .0930 & .1566 & - & - & - \\ \hline
\multirow{9}{30pt}{\centering CNN \\ (15 layers)}& \multirow{4}{24pt}{\centering Learnt (50\% data)}&  $f_{\mathrm{quad}}$ & .0867 & \emph{.2174} & .0053 & .0587 & .1753 & .0584 & .1668 & .0538 & .1662 & .0014 & .1653 & .1688 \\
& & $f_{\mathrm{lambda}}$ & .1217 & .2707 & .0304 & .0494 & .1661 & .0506 & \emph{.1618} & .0417 & .1639 & .0015 & .1622 & .1688 \\
& & $f_{\mathrm{classic}}$ & .0782 & \textbf{.1954} & .0007 & .0667 & .1783 & .0652 & .1686 & .0594 & .1692 & .0013 & .1674 & .1688 \\
& & $f_{\mathrm{bbb}}$ & .6069 & .7066 & .7908 & .0287 & .1073 & .0468 & \textbf{.1553} & .0412 & .1530 & .0012 & .1517 &  .1688 \\  \cline{2-15}
& \multirow{4}{24pt}{\centering Learnt (70\% data)}&  $f_{\mathrm{quad}}$ & .0756 &  \emph{.1806} &  .0028 &  .0559 &  .1513 &  .0559 &  .1463 &  .0391 &  .1469 &  .0016 &  .1449 &  .1490 \\
& & $f_{\mathrm{lambda}}$ & .0922 &  .2121 &  .0133 &  .0486 &  .1477 &  .0500 &  \emph{.1437} &  .0507 &  .1449 &  .0012 &  .1438 &  .1490 \\
& & $f_{\mathrm{classic}}$ &.0703 &  \textbf{.1667} &  .0003 &  .0622 &  .1548 &  .0615 &  .1475 &  .0551 &  .1480 &  .0010 &  .1476 &  .1490 \\
& & $f_{\mathrm{bbb}}$ &.4481 &  .5572 &  .4795 &  .0259 &  .0947 &  .0455 &  \textbf{.1413} &  .0395 &  .1405 &  .0008 &  .1409 &  .1490 \\ \cline{2-15}
& - & $f_{\mathrm{erm}}$ & - & - & - & .0208 & .0339 & - & - & .0957 & .1413 & - & - & - \\
\hline
\end{tabular}
% \end{center}
\caption{Training and test set results on CIFAR-10 using Gaussian distributions over weights. 
The table includes results for three deep CNN architectures (with 9, 13, and 15 layers, respectively) and data-dependent PAC-Bayes priors which are obtained via empirical risk minimisation for learning the prior mean using two percentages of the data (50\% and 70\%, corresponding to 25.000 and 35.000 examples respectively). For the stochastic predictor, the best risk certificate and test set error are highlighted in bold face, and second best are highlighted in italics.} 
\label{tab:cifar}
\end{table}

\begin{figure}[ht]
\centering
\includegraphics[width=0.95\textwidth]{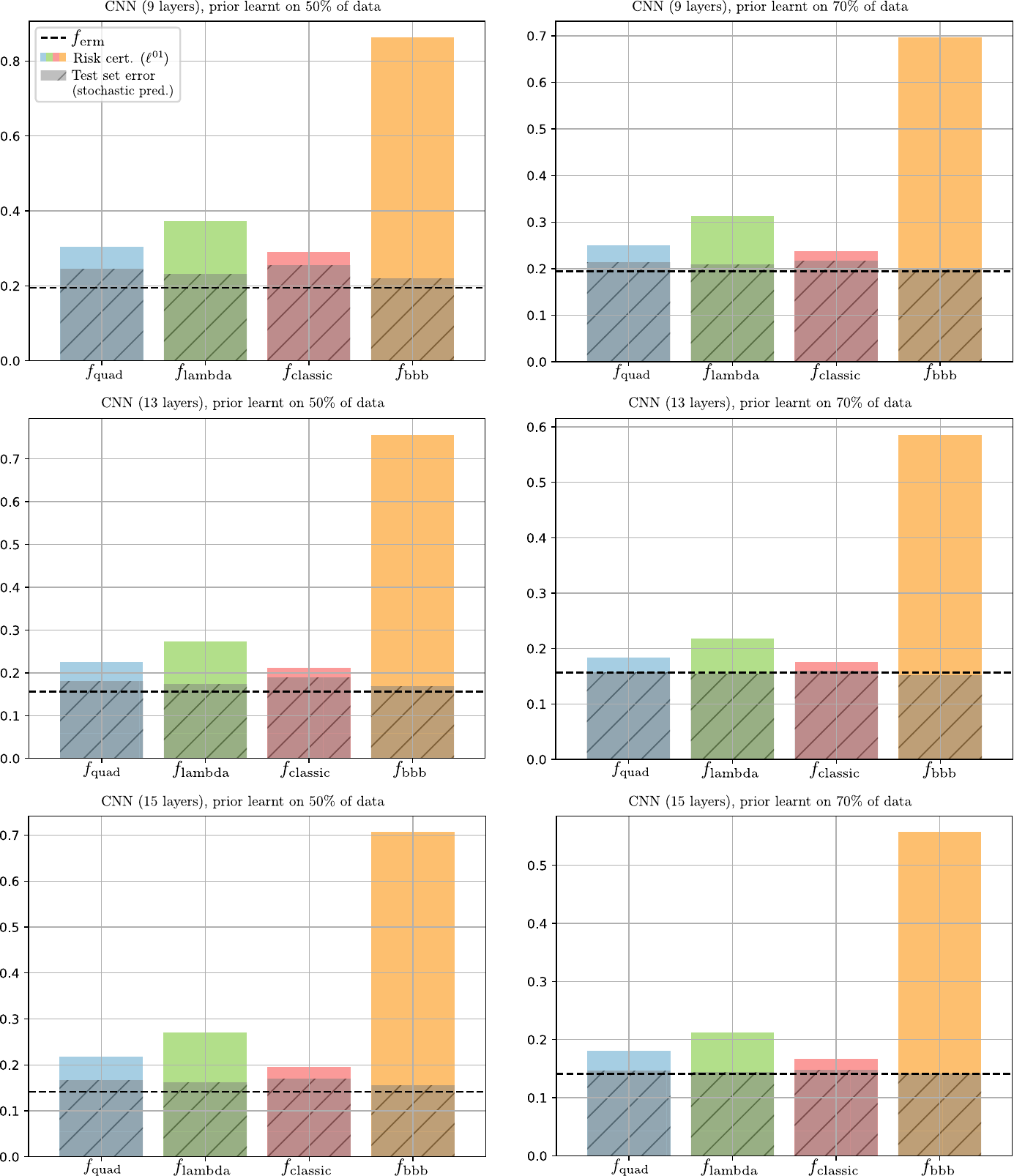}
\caption{
% Bar plots of results achieved on CIFAR-10 for 3 different network architectures and two data-dependent priors (learnt using 50\% and 70\% of the data).
Tightness of the risk certificates on CIFAR-10 for 3 different network architectures and two data-dependent priors (learnt using 50\% and 70\% of the data).
}
\label{fig:cifarbar}
\end{figure}

\begin{figure}[t]
\centering
\includegraphics[width=9cm]{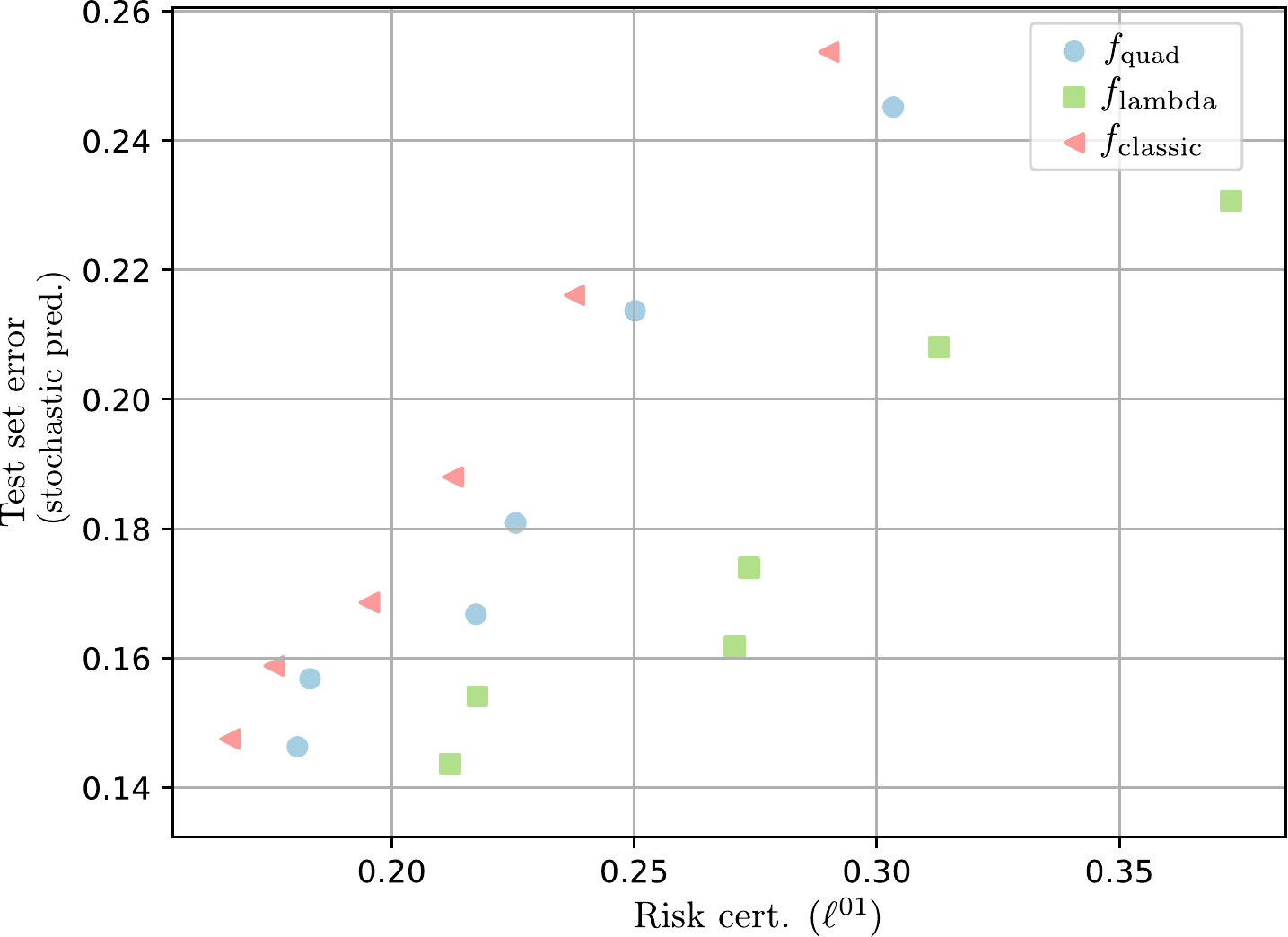}
\caption{Scatter plot of the results obtained for CIFAR-10 using three different training objectives. 
The x-axis shows values of the risk certificate 
(under $\ell^{01}$ loss),
and the y-axis shows the test set error rates, 
achieved by the stochastic classifier.}
\label{fig:cifarplotall}
\end{figure}

We evaluate now our training objectives on CIFAR-10 using deep CNN architectures. Note that this is a much larger scale experiment than the ones presented before (15 layers with learnable parameters vs 4). As far as we know, we are the first to evaluate PAC-Bayes inspired training objectives in such deep architectures. The results are presented in \cref{tab:cifar} and \figurename{ \ref{fig:cifarbar}} for three architectures (with 9, 13 and 15 layers, with around 6M, 10M and 13M parameters, respectively). Note, however, that the number of parameters is doubled for our probabilistic neural networks. We also experiment with using different amount of data for learning the prior: 50\% and 70\%, leaving respectively 25.000 and 15.000 examples to evaluate the bound. The conclusions are as follows: i) In this case, the improvements brought by learning the posterior through PBB with respect to the prior are much better and generally consistent across all experiments (e.g. 2 points in test 0-1 error for $f_{\mathrm{lambda}}$ when using 50\% of the data for learning the prior). ii) Risk certificates are also non-vacuous and tight (although less than for MNIST). iii) We validate again that $f_{\mathrm{lambda}}$ shows better test performance but less tight risk certificates. iv) In this case, however, $f_{\mathrm{classic}}$ and $f_{\mathrm{quad}}$ seem much closer in terms of performance and tightness. In some cases, $f_{\mathrm{classic}}$ provides slightly tighter bounds, but also often worse test performance. The tighter bounds can be explained by our findings with the Pinsker inequality, which makes $f_{\mathrm{classic}}$ tighter when true loss is more than 0.25. This observation can be seen clearly in \figurename{ \ref{fig:cifarplotall}}.
v) Obtained results with 15 layers are competitive, achieving similar performance than those reported for VGG-16 \citep{Simonyan15} (deep network proposed for CIFAR-10 with comparable architecture to the one tested with only fully connected and convolutional layers). 
vi) The results indicate that 50\% of the training data is not enough in this data set to build a competitive prior and this influences the test performance and the risk certificates. The results with 70\% of the data are, however, very close to those achieved by ERM across all three architectures. vii) Similarly than with the rest of the experiments, a major difference can be seen when comparing the risk certificate achieved by $f_{\mathrm{bbb}}$ with the risk certificate achieved by PAC-Bayes inspired training objectives. viii) Finally, it is noteworthy how the KL gets generally smaller as we move to deeper architectures (specially from 9 to 13 layers), which is an interesting observation, as there are many more parameters used in the computation of the KL. This indicates that the posterior in deeper architectures stays much closer to the prior. We believe this may be because in a higher-dimensional weight space, the weight updates have a smaller euclidean norms, hence the smaller KL.

Note that more competitive and deeper neural baselines exist for CIFAR-10 nowadays. However, those deeper architectures often require of more advanced training strategies such as batch norm, data augmentation, cyclical learning rates, weight decay, etc. 
In our experiments, we decided to keep the training strategy as simple as possible,
in order to focus on the ability of our training objectives alone to give good predictors and, more importantly, risk certificates with tight values.
It is noteworthy that our training objectives are able to achieve this with a simple training strategy, and we leave the exploration of all the available training choices as future work.

\subsection{Additional Miscellaneous Experiments}

In this section we discuss four interesting observations from our experiments, which we believe mark promising future research directions.   

\begin{figure}[t]
\centering
\includegraphics[width=0.9\textwidth]{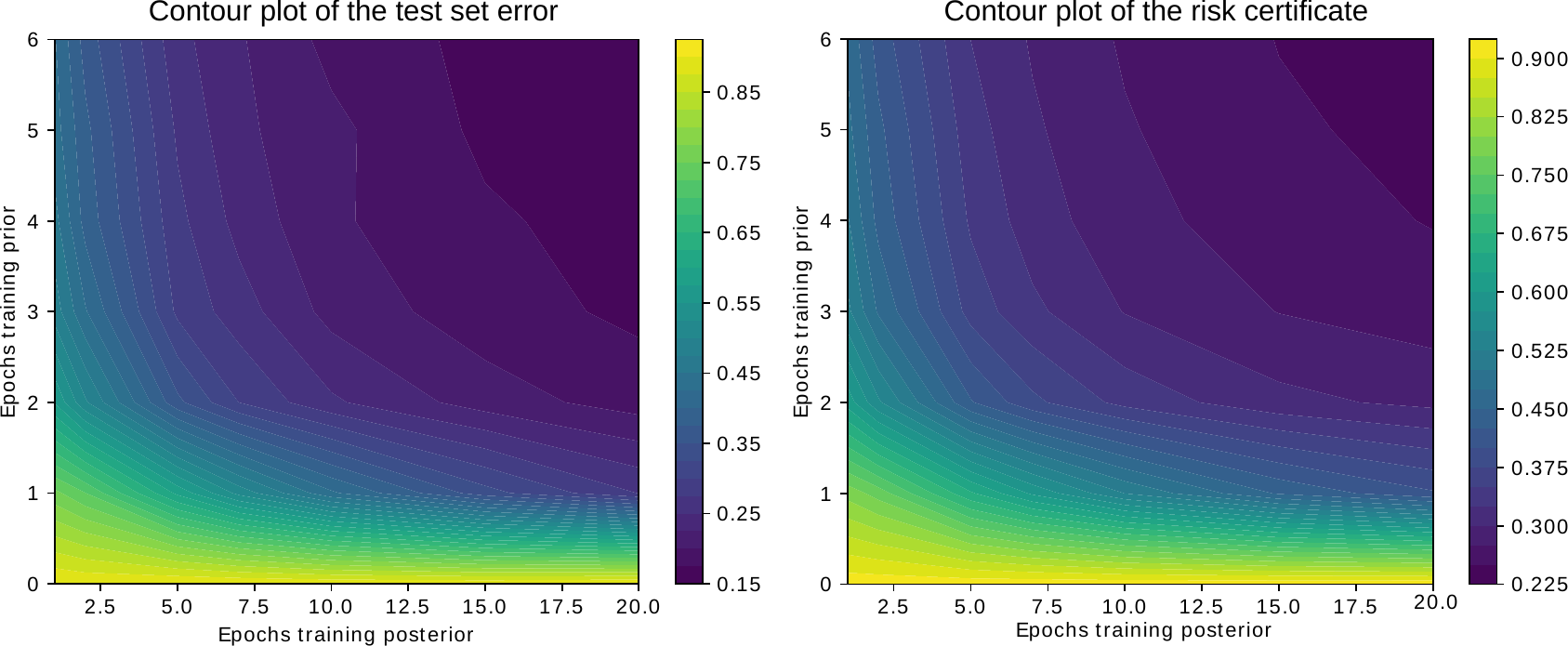}
\caption{Contour plots of the test set error and risk certificate (under $\ell_{01}$) after different training epochs learning the prior and posterior and initial scale hyper-parameters value $\sigma_0=0.1$ for the prior.}
\label{fig:contours}
\end{figure}

\begin{figure}[ht]
\centering
\includegraphics[width=0.6\textwidth]{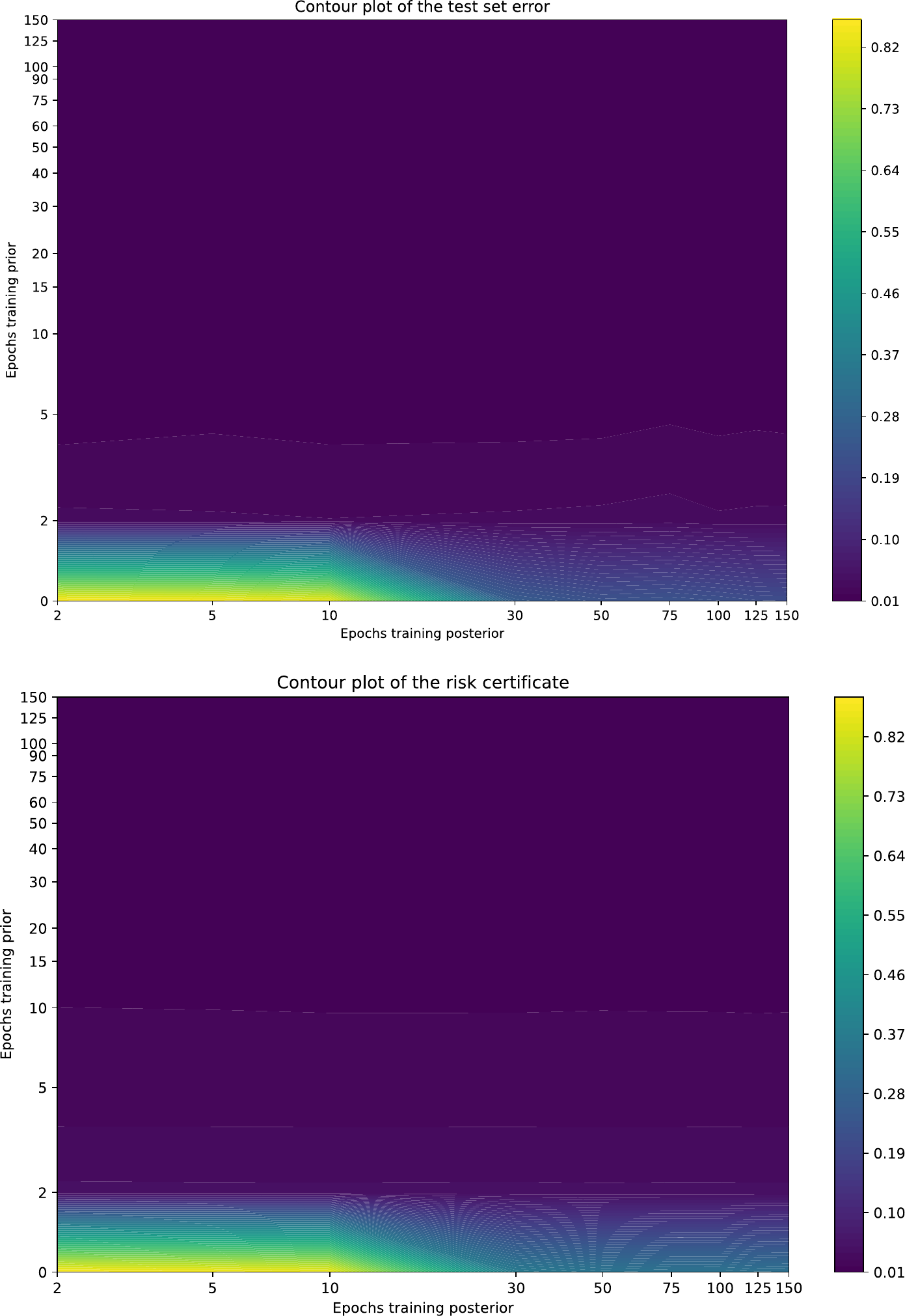}
\caption{Contour plots of the test set error and risk certificate (under $\ell_{01}$) after different training epochs learning the prior and posterior. Dropout is used when learning the prior. Note that training the posterior for a large number of epochs does not worsen the test set error or the risk certificate.}
\label{fig:contourslonger}
\end{figure}

First, we present a plot of the performance obtained when using different training epochs to learn the prior and posterior.
\figurename{ \ref{fig:contours}} and \ref{fig:contourslonger} show a contour plot of the loss and risk certificate when training the prior and the posterior for different epochs (e.g. to check the effect of training the posterior with an under-fitted prior). These plots have been generated using the FCN architecture on MNIST with Gaussian distributions over weights. Similar results are obtained for the CNN architecture. Note that for the sake of visualisation in \figurename{ \ref{fig:contours}} we are plotting much less epochs that those used to generate the final results (in this case up to 20, whereas the rest of reported results were with 100 epochs) so the reported test set errors and risk certificates in this plot differ from those previously reported. \figurename{ \ref{fig:contours}} shows that both training the prior and the posterior are crucial to improve the final loss and risk certificates, as the best loss and risk certificate values are found in the top right corner of the plot. 
The plot also shows that if the prior is under-fitted (e.g. if trained for only one or two epochs), then the final predictor can still be much improved with more training epochs for the posterior. However, a more adequate prior means that less epochs are needed to reach a reasonable posterior. Nonetheless, this is less apparent if the prior is not learnt (represented here as a training of 0 epochs, i.e. a random prior), in which case learning the posterior for longer does not seem to reach such competitive posteriors, which demonstrates the usefulness of data-dependent priors for obtaining tight risk certificates.
In this experiment depicted in \figurename{ \ref{fig:contours}} and \ref{fig:contourslonger}, we also noted that only a few epochs of training the prior are enough to reach competitive posteriors and that learning the posterior for much longer (e.g. 1000 epochs) does not lead to overfitting, which reinforces the finding of \cite{blundell2015weight} that the KL term act as a regulariser. Specifically, this can be seen in \figurename{ \ref{fig:contourslonger}}, which  shows that training the posterior for a large number of epochs does not worsen the test set error and the risk certificate.
There are still small scale differences (of up to 1\%) in risk certificate and test set error for the dark blue colour region, but these can not be visually seen because of the scale of the colour legend. 
However, the important observation is that the differences are small across the dark blue region (if there were significant differences within this region, then that would be an evidence of overfitting).
This is, however, opposed to what we observe when training the prior through empirical risk minimisation, since the prior overfits easily in that case, which is why we had to learn the priors using dropout in all our experiments.

\smallskip

\begin{figure}[ht]
\centering
\includegraphics[width=0.95\textwidth]{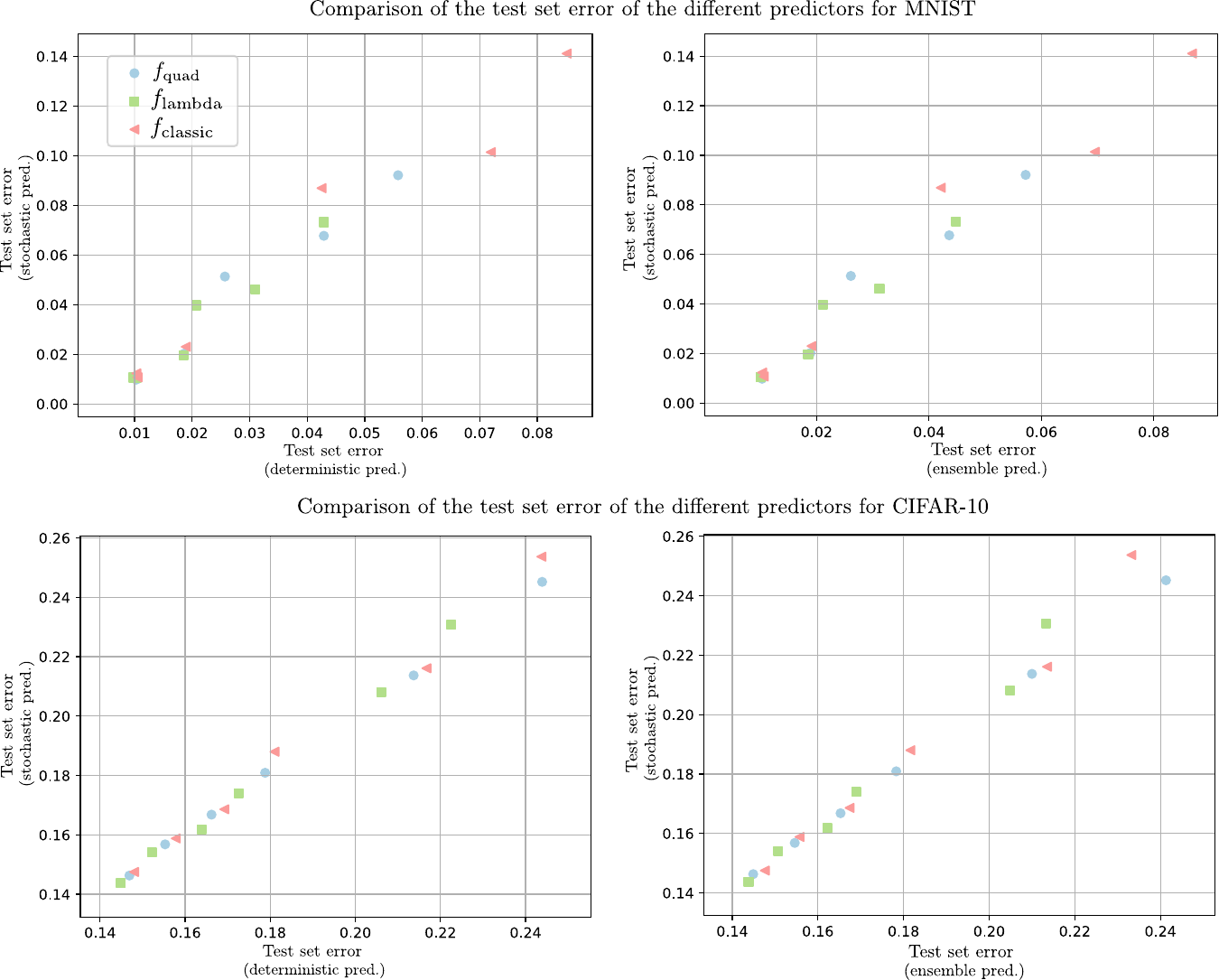}
\caption{Representation of the results achieved by the different predictors that were studied (stochastic, deterministic, and ensemble). }
\label{fig:predictors}
\end{figure}

Next, we compare the test set performance of the different predictors considered in this work (stochastic, deterministic and ensemble). The results for MNIST and CIFAR-10 are depicted in \figurename{ \ref{fig:predictors}}. One can appreciate a very clear linear relationship between predictors. In the case of CIFAR-10 the results are similar across all predictors, whereas for MNIST the stochastic predictor obtains significantly worse results (see differences in scales of x and y axes). In the case of CIFAR-10 this may hint that 
our training strategy finds a solution within a large region of comparably good solutions, so that weight randomisation does not affect significantly the test performance of the classifier.
We plan to explore this interesting phenomenon in future work.

\smallskip

\begin{figure}[ht!]
\centering
\includegraphics[width=0.8\textwidth]{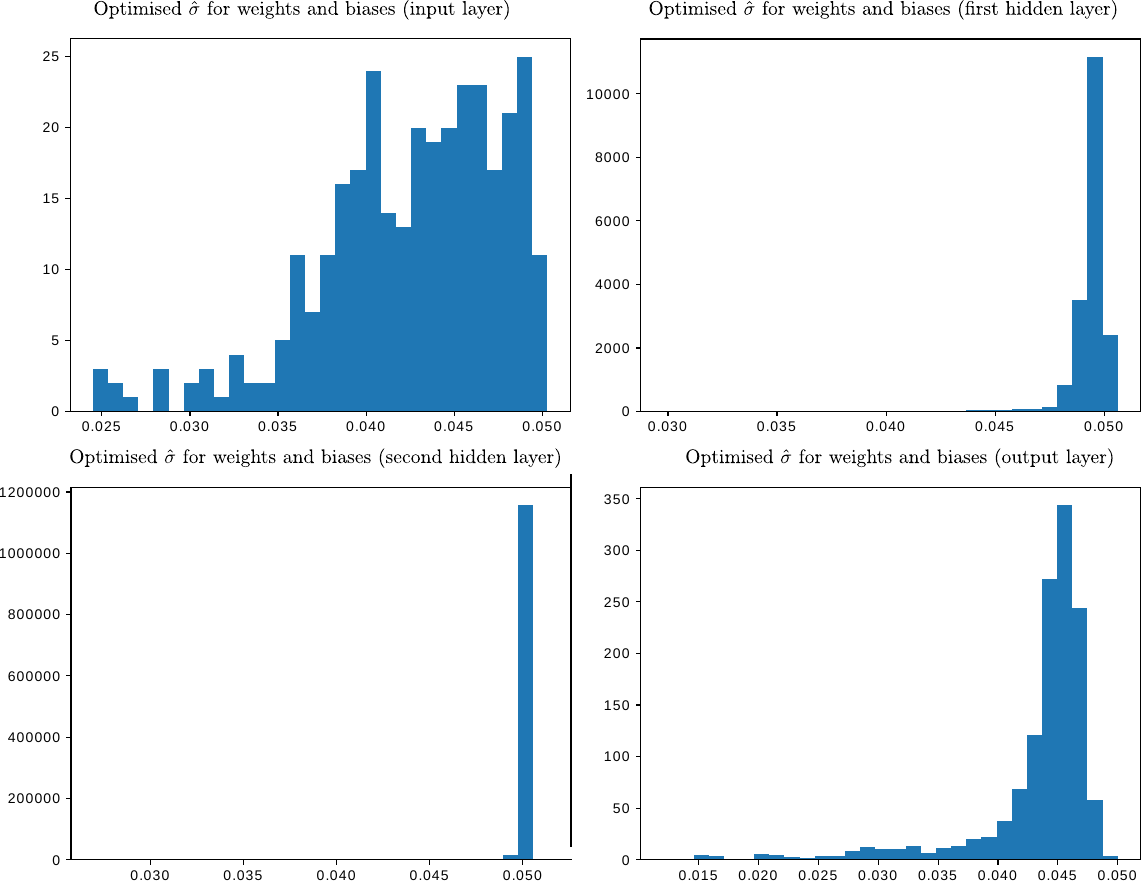}
\caption{Histograms of the scale parameters for the Gaussian distribution at the end of the optimisation for the different layers of the CNN architecture on MNIST. All scale parameters were initialised to 0.05, i.e. $\sigma_0 = 0.05$ for all coordinates, and $\hat{\sigma}$ is the scale parameter value of the final output of training.}
\label{fig:histogramsigmas}
\end{figure}

Thirdly, in \figurename{ \ref{fig:histogramsigmas}} we show a histogram of the final scale parameters $\hat{\sigma}$ (i.e. standard deviation) for the Gaussian posterior distribution (both weights and biases). The plot shows that the optimisation changes the scale of different weights and biases, reducing specially those associated to the input and output layer. We think it is worth to experiment with different scale initialisations per layer in future work, as well as different covariance structures for the weight and bias distributions.

\begin{figure}[ht!]
\centering
\includegraphics[width=0.4\textwidth]{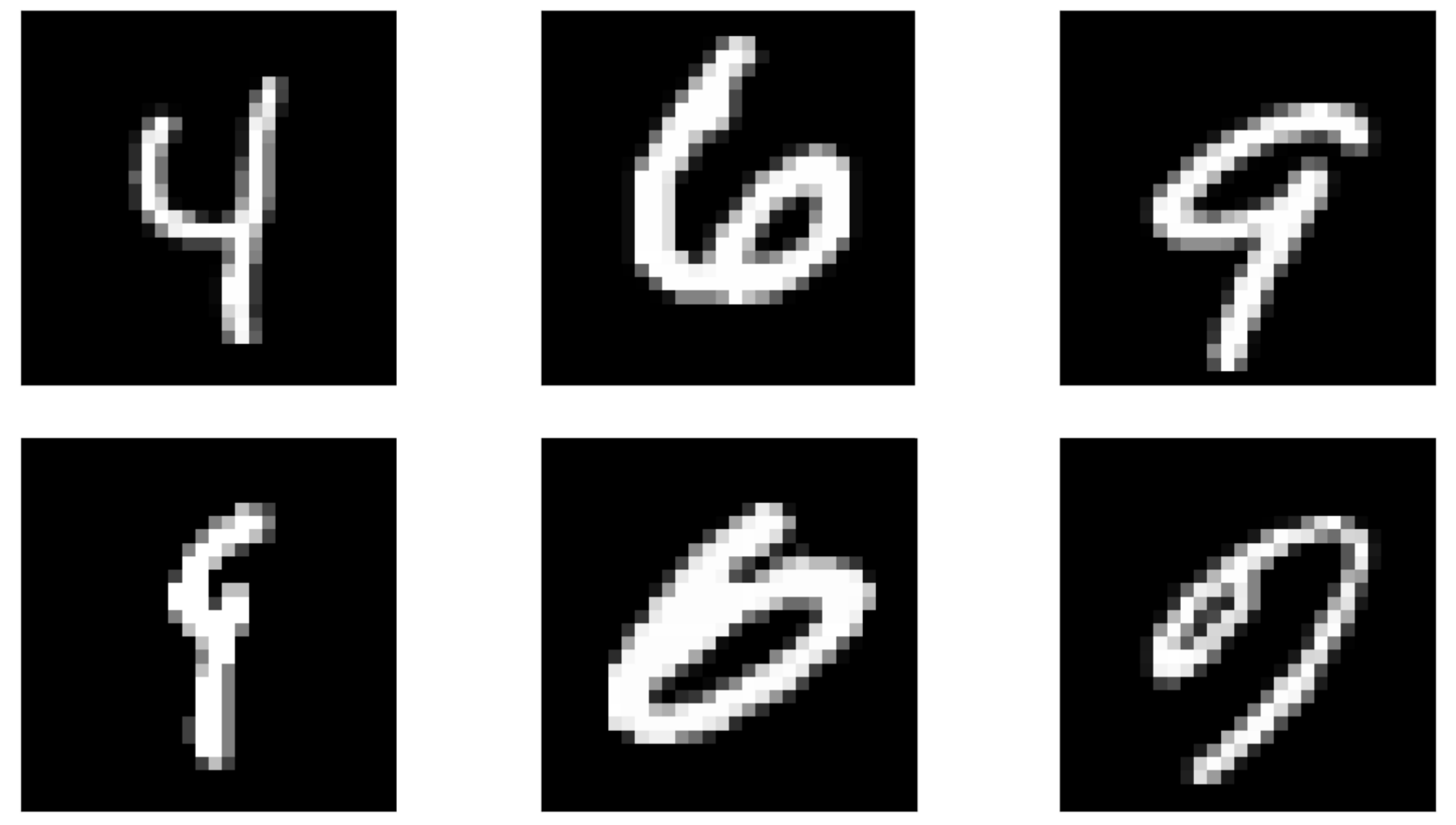}
\caption{Representation of the test set digits for three classes (4, 6 and 9) in which the ensemble predictor is most certain/uncertain. The top row shows the digits with minimum uncertainty (all 100 members of the ensemble agree in the prediction). The bottom row shows the digits with highest uncertainty.}
\label{fig:uncertainty}
\end{figure}

\smallskip

Finally, we aim to validate the use of the learnt posterior for uncertainty quantification. To do so, we use the ensemble predictor (100 members) using the CNN architecture in MNIST. Each member of the ensemble  is a sample from the posterior. We define uncertainty as the number of members of the ensemble that disagree in the prediction.\footnote{
Similar measures of disagreement have been used in the literature on majority vote classifiers, see e.g. \cite{Lacasse-etal2006,germain-etal2015,masegosa2020second}; and
in some related literature on domain adaptation, e.g. \cite{germain2013_DA,germain2016_DA,germain2020_DA}.
} 
\figurename{ \ref{fig:uncertainty}} shows the test set digits for which the ensemble is most certain (top row) and uncertain (bottom row). It can be seen that the most uncertain digits indeed look unusual and could even confuse a human, whereas the most certain digits are easily identifiable as 4, 6 and 9. 
We believe that this simple visual experiment may indicate that there is promise in probabilistic neural networks trained by %
PBB objectives being of use for uncertainty quantification. However, more experiments in this direction are needed.

\subsection{Further Discussion}

We now discuss further the probabilistic neural network models studied in this paper, with a focus on their practical usefulness. 
We have demonstrated that 
the randomised predictors learnt by PBB come with a tight performance guarantee that is valid at population level, 
and is evaluated on a subset of the data used to train the PAC-Bayes posterior, i.e. evaluation of our certificates does not require a held out test set.
We have observed that our methods show promise for self-certified learning, which is a data-efficient principle,
and also
 shown that the same bound used for post-training evaluation of the risk certificate is  useful for model selection.
Practitioners may want to consider all these favourable properties.

However, probabilistic neural networks have additional advantages over their standard point estimator counterparts. 
The results of \citet{blundell2015weight} have shown that probabilistic neural networks enable an intuitive and principled implementation of uncertainty quantification and classification reject options (e.g. allow the model to say ``I don't know'' when the classification uncertainty for a new example is higher than a certain threshold).
Similarly, we have also shown the use of our models for uncertainty quantification in a very simple experiment with the ensemble predictor. 
This is just one example of the advantages of probabilistic neural network models (distributions over weights) compared to using point estimator models (fixed weights), but these models have shown promise towards many other goals, such as model pruning/distillation.
\citet{blundell2015weight} also showed that learning a weight distribution by minimising the empirical loss while constraining its KL divergence to a prior
gives similar results to %
implicit regularisation schemes (such as dropout). 
Similarly, in the experiments 
with our training objectives we have seen that overfitting was only an issue while learning the prior through ERM, but not during the posterior learning phase (as demonstrated by \figurename{ \ref{fig:contourslonger}}).

Even though we have not experimented exhaustively with all of the cases described above, we hypothesise that all these advantages extend to probabilistic neural networks learnt by PAC-Bayes inspired objectives. 
This may make the use of stochastic classifiers with PAC-Bayes bounds more desirable than point estimator models with a PAC bound.
This is also notwithstanding the tightness of the former, in contrast with the latter, which are known to be notoriously vacuous for the kinds of models studied in our experiments. All of these hypotheses should be validated thoroughly in future work.

\section{Conclusion and Future Work}
\label{s:conclude}

In this paper we explored `PAC-Bayes with Backprop' (PBB) methods to train probabilistic neural networks with different weight distributions, priors and network architectures. 
The take-home message is that the training methods presented in this paper are derived from sound theoretical principles and provide a simple strategy that comes with a performance guarantee that is valid at population level, i.e. valid for any unseen data from the same distribution as the training data. This is an improvement over methods derived heuristically rather than from theoretically justified arguments, and over methods that do not include a risk certificate valid on unseen examples. Additionally, we empirically demonstrate the usefulness of data-dependent priors for achieving competitive test set performance and, importantly, for computing risk certificates with tight values. 

The results of our experiments on MNIST and CIFAR-10 have showed that these PBB objectives give predictors with competitive test set performance and with  non-vacuous risk certificates that significantly improve previous results and can be used not only for guiding the learning algorithm and certifying the risk but also for model selection.
This shows that PBB methods are promising examples of self-certified learning, since the values of the risk certificates output by these training methods are tight, i.e. close to the values of the test set error estimates. 
In particular, our results in MNIST with a small convolutional neural network (2 hidden layers) achieve 1\% test set error and a risk certificate of 1.5\%.
We also evaluated our training objectives on large convolutional neural networks (up to 15 layers and around 13M parameters) with CIFAR-10.
These results also showed risk certificates with tight values (18\% of risk certificate for a stochastic predictor that achieves 14.6\% of test set error).
Note that to claim that self-certified learning is achieved would require testing a given training method across a wide range or data sets and architectures (so as to experimentally validate the claim), or theoretically characterising the problems on which a given learning method is guaranteed to produce tight risk certificates.

In future work we plan to test different covariance structures for the weight distribution and validate a more extensive list of choices for the weight distributions across a larger list of data sets. We also plan to experiment how to approach the well-known dominance of the KL term in the optimisation of these objectives. Data-dependent priors seem like a promising avenue to do so. We also plan to explore deeper architectures. Finally, we plan to study risk certificates for the ensemble predictor. 
We also plan to study different ensemble methods, for instance the one that \citet{thiemann-etal2017} used with SVMs looks promising, it would be interesting to explore such method (and others) with neural networks.

\begin{acks}
We warmly thank the anonymous reviewers and the action editor for their valuable feedback, which helped us to improve the paper greatly. 

For comments on various early parts of this work we warmly thank Yevgeny Seldin, Charles Blundell, Andriy Mnih, Nando de Freitas, Razvan Pascanu, Benjamin Guedj,
and Pascal Germain. 
An early version of this work was presented at the NeurIPS 2020 workshop `Beyond Backpropagation' \citep{perez-ortiz2020towards}.

We warmly acknowledge the AI Centre at University College London, and DeepMind, for providing friendly and stimulating work environments. 

Mar\'ia P\'erez-Ortiz and John Shawe-Taylor gratefully acknowledge support and funding from the U.S. Army Research Laboratory and the U. S. Army Research Office, and by the U.K. Ministry of Defence and the U.K. Engineering and Physical Sciences Research Council (EPSRC) under grant number EP/R013616/1.

Omar Rivasplata gratefully acknowledges sponsorship from DeepMind  for carrying out research studies in machine learning at University College London. This work was done while Omar was a research scientist intern at DeepMind.

Csaba Szepesv{\'a}ri gratefully acknowledges funding from the Canada CIFAR AI Chairs Program, the Alberta Machine Intelligence Institute (Amii), and the Natural Sciences and Engineering Research Council (NSERC) of Canada.
\end{acks}

% \newpage
% \appendix

\end{document}